\documentclass[12pt]{colt2018} 

\title[What Doubling Tricks Can and Can't Do for Multi-Armed Bandits]{What Doubling Tricks Can and Can't Do for Multi-Armed Bandits}

\usepackage{macrosArticle}
\usepackage{macrosText}
\usepackage{times}
\usepackage{array}
\usepackage{float}

\coltauthor{\Name{Lilian Besson}\textsuperscript{$\dagger$} \Email{{Lilian}{.}{Besson}{@}{CentraleSupelec}{.}{fr}} \\
    \addr CentraleSup\'elec (campus of Rennes), IETR, SCEE Team,\\
    Avenue de la Boulaie -- CS $47601$, F-$35576$ Cesson-S\'evign\'e, France
    \AND
    \Name{Emilie Kaufmann} \Email{{emilie}{.}{kaufmann}{@}{univ}{-}{lille1}{.}{fr}} \\
    \addr CNRS \& Universit\'e de Lille, Inria SequeL team\\
    UMR $9189$ -- CRIStAL,  F-$59000$ Lille, France
}

\begin{document}

\maketitle

\begin{abstract}
    An online reinforcement learning algorithm is \emph{anytime} if it does not need to know in advance the horizon $T$ of the experiment.
    A well-known technique to obtain an anytime algorithm from any non-anytime algorithm is the ``Doubling Trick''.
    In the context of adversarial or stochastic multi-armed bandits,
    the performance of an algorithm is measured by its regret,
    and we study two families of sequences of growing horizons (geometric and exponential)
    to generalize previously known results that certain doubling tricks can be used to conserve certain regret bounds.
    In a broad setting, we prove that a geometric doubling trick can be used to conserve (minimax) bounds in $R_T = \cO(\sqrt{T})$ but \emph{cannot} conserve (distribution-dependent) bounds in $R_T = \cO(\log T)$.
    We give insights as to why exponential doubling tricks may be better, as they conserve bounds in $R_T = \cO(\log T)$, and are close to conserving bounds in $R_T = \cO(\sqrt{T})$.
    %

\end{abstract}

\begin{keywords}
    Multi-Armed Bandits;
    Anytime Algorithms;
    Sequential Learning;
    Doubling Trick.
\end{keywords}

\section{Introduction}

Multi-Armed Bandit (MAB) problems are well-studied sequential decision making problems in which an agent repeatedly chooses an action (the ``arm'' of a one-armed bandit) in order to maximize some total reward \citep{Robbins52,LaiRobbins85}.
Initial motivation for their study came from the modeling of clinical trials, as early as 1933 with the seminal work of \cite{Thompson33}.
In this example, arms correspond to different treatments with unknown, random effect.
Since then, MAB models have been proved useful for many more applications, that range from
cognitive radio \citep{Jouini09} to
online content optimization (\eg, news article recommendation \citep{Li10contextual}, online advertising \citep{LiChapelle11}, A/B Testing \citep{Kaufmann14,Jamieson17ABTest}),
or portfolio optimization \citep{Sani2012risk}.

While the number of patients involved in a clinical study (and thus the number of treatments to select) is often decided in advance, in other contexts the total number of decisions to make (the horizon $T$) is unknown.
It may correspond to the total number of visitors of a website optimizing its displays for a certain period of time, or to the number of attempted communications in a smart radio device.
In such cases, it is thus crucial to devise \emph{anytime algorithms}, that is algorithms that do no rely on the knowledge of this horizon $T$ to sequentially select arms.
A general way to turn any base algorithm into an anytime algorithm is the use of the so-called Doubling Trick, first proposed by \cite{auer1995gambling}, that consists in repeatedly running the base algorithm with increasing horizons.
Motivated by the frequent use of this technique and the absence of a generic study of its effect on the algorithm's efficiency,
this paper investigates in details two families of doubling sequences (geometric and exponential), and shows that the former should be avoided for stochastic problems.

More formally, a MAB model is a set of $K$ arms, each arm $k$ being associated to a (unknown) \emph{reward stream} $(Y_{k,t})_{t\in\N}$.
Fix $T$ a finite (possibly unknown) horizon.
At each time step $t \in \{1,\dots,T\}$ an agent selects an arm $A(t) \in \{1,\dots,K\}$ and receives as a reward the current value of the associated reward stream, $r(t) := Y_{A(t), t}$.
The agent's decision strategy (or \emph{bandit algorithm})
$\cA_T := (A(t), t\in \{1,\dots,T\})$ is such that $A(t)$ can only rely on the past observations
$A(1),r(1),\dots,A(t-1),r(t-1)$,
on external randomness and (possibly) on the knowledge of the horizon $T$.
The objective of the agent is to find an algorithm $\cA$ that maximizes the expected cumulated rewards, where the expectation is taken over the possible randomness used by the algorithm and the possible randomness in the generation of the rewards stream.
In the oblivious case, in which the reward streams are independent of the algorithm's choice, this is equivalent to minimizing the \emph{regret}, defined as
\begin{equation}\label{eq:regret}
    R_T(\cA_T) := \max_{k \in \{1,\dots,K\}}
    \E\left[\sum_{t=1}^{T} \left(Y_{k,t} - Y_{A(t),t}\right)\right].
\end{equation}

This quantity, referred to as \emph{pseudo-regret} in \cite{Bubeck12}, quantifies the difference between the expected cumulated reward of the best fixed action, and that of the strategy $\cA_T$.
For the general adversarial bandit problem \citep{Auer02NonStochastic}, in which the rewards streams are arbitrary (picked by an adversary), a \emph{worst-case} lower bound has been given.
It says that for every algorithm, there exists (stochastic) reward streams such that the regret is larger than $(1/20)\sqrt{KT}$
\citep{Auer02NonStochastic}.
Besides, the \ExpThree{} algorithm has been shown to have a regret of order $\sqrt{KT\log(K)}$.

Much smaller regret may be obtained in \emph{stochastic} MAB models, in which the reward stream from each arm $k$ is assumed to be \iid, from some (unknown) distribution $\nu_k$, with mean $\mu_k$.
In that case, various algorithms have been proposed with \emph{problem-dependent} regret upper bounds of the form $C(\bm\nu) \log(T),$ where $C(\bm\nu)$ is a constant that only depend on the arms distributions.
Different assumptions on the arms distributions lead to different problem-dependent constants.
In particular, under some parametric assumptions (\eg, Gaussian distributions, exponential families), \emph{asymptotically optimal} algorithms have been proposed and analyzed (\eg, \KLUCB{} \citep{KLUCBJournal} or Thompson sampling \citep{AgrawalGoyal11,Kaufmann12Thompson}), for which the constant $C(\bm\nu)$ obtained in the regret upper bound matches exactly that of the lower bound given by \cite{LaiRobbins85}.
Under the non-parametric assumption that the $\nu_k$ are bounded in $[0,1]$, the regret of the \UCBone{} algorithm \citep{Auer02} is of the above form with $C(\bm\nu) = 8 \times \sum_{k : \mu_k > \mu^*} (\mu^* - \mu_k)^{-1}$, where $\mu^* = \max_k \mu_k$ is the mean of the best arm.
Like in this last example, all the available constants $C(\bm\nu)$ become very large on ``hard'' instances, in which some arms are very close to the best arm.
On such instances,  $C(\bm\nu)\log(T)$ may be much larger than the worst-case $(1/20)\sqrt{KT}$, and distribution-independent guarantees may actually be preferred.

The \MOSS{} algorithm, proposed by \cite{Audibert2009minimax}, is the first stochastic bandit algorithm
to enjoy a problem-dependent logarithmic regret
and to be optimal in a \emph{minimax} sense, as its regret is proved to be upper bounded by $\sqrt{KT}$, for bandit models with rewards in $[0,1]$.
However the corresponding constant $C(\bm \nu)$ is proportional to $K/\Delta_{\min}$, where $\Delta_{\min} = \min_{k} (\mu^*-\mu_k)$ is the minimal gap, which worsen the constant of \UCBone{}.
Another drawback of \MOSS{} is that it is \emph{not} anytime.
These two shortcoming have been overcame recently in two different works.
On the one hand, the \MOSSAnytime{} algorithm \citep{Degenne16} is minimax optimal and anytime, but its problem-dependent regret does not improve that of \MOSS.
On the other hand, the \KLUCBpp{} algorithm \citep{Menard17} is simultaneously minimax optimal and asymptotically optimal (\ie, it has the best problem-dependent constant $C(\bm\nu)$), but it is not anytime. A natural question is thus to know whether a Doubling Trick could overcome this limitation.

This question is the starting point of our comprehensive study of the Doubling Trick: can a single Doubling Trick be used to preserve both problem-dependent (logarithmic) regret and minimax (square-root) regret?
We answer this question in the negative, by showing that two different types of Doubling Trick may actually be needed.
In this paper, we investigate how algorithms enjoying regret guarantees of the generic form
\begin{equation}\label{eq:genericWithLogT}
    \forall T \geq 1,\;\;\;\; R_T(\cA_T) \leq c \; T^{\gamma} (\log(T))^{\delta} + o( T^{\gamma} \left(\log(T))^{\delta} \right)
\end{equation}
may be turned into an anytime algorithm enjoying \emph{similar} regret guarantees with an appropriate Doubling Trick.
This does not come for free, and we exhibit a ``price of Doubling Trick'', that is a constant factor larger than $1$, referred to as a \emph{constant manipulative loss}.

The rest of the paper is organized as follows.
The Doubling Trick is formally defined in Section~\ref{sec:genericAlgo}, along with a generic tool for its analysis.
In Section~\ref{sec:geometricDoublingTrick}, we present upper and lower bounds on the regret of algorithms to which a geometric Doubling Trick is applied.
Section~\ref{sec:exponentialDoublingTrick} investigates regret guarantees that can be obtained for a ``faster'' exponential Doubling Trick.
Experimental results are then reported in Section~\ref{sec:experiments}.
Complementary elements of proofs are deferred to the appendix.

\section{Doubling Tricks }\label{sec:genericAlgo}

The Doubling Trick, denoted by \DT, is a general procedure to convert a (possibly non-anytime) algorithm into an anytime algorithm.
It is formally stated below as Algorithm~\ref{algo:DTr} and depends on a non-decreasing diverging \emph{doubling sequence} $(T_i)_{i\in \N}$
(\ie, $T_i\to\infty$ for $i\to\infty$).
\DT{} fully restarts the underlying algorithm $\cA$ at the beginning of each new sequence (at $t = T_i + 1$), and run this algorithm on a sequence of length ($T_i - T_{i-1}$).


%

%
%




\vspace*{5pt}  
\begin{algorithm}[H]
        \LinesNumbered  
        \DontPrintSemicolon
        \KwIn{Bandit algorithm $\cA$, and doubling sequence $(T_i)_{i\in\N}$.}
        \BlankLine
        Let $i = 0$, and initialize algorithm $\cA^{(0)} = \cA_{T_0}$.\;
        \For{$t = 1, \dots, T-1$}{
            \If(\tcp*[f]{Next horizon $T_{i+1}$ from the sequence}){
                $t > T_i$
            }{
                Let $i = i + 1$.\;
                Initialize algorithm $\cA^{(i)} = \cA_{T_i - T_{i-1}}$.
                \tcp*[f]{Full restart}
            }
            Play with $\cA^{(i)}$: play arm $A'(t) := A^{(i)}(t - T_i)$, observe reward $r(t) = Y_{A'(t), t}$. \nllabel{line:internalTimer_DTr}
        }
        \caption{The Generic Doubling Trick Algorithm, $\cA' = \DTr(\cA, (T_i)_{i\in\N})$.}
\label{algo:DTr}
\end{algorithm}
\vspace*{5pt}  

%

\paragraph{Related work.}
The Doubling Trick is a well known idea in online learning, that can be traced back to \cite{auer1995gambling}.
In the literature, the term Doubling Trick usually refers to the geometric sequence $T_i = 2^i$, in which the horizon is actually \emph{doubling}, that was popularized by
\cite{CesaLugosi06} in the context of adversarial bandits. Specific doubling tricks have also been used for stochastic bandits, for example in the work of \cite{Auer10}, which uses
the doubling sequence $T_i = 2^{2^{i}}$ to turn the \UCBR{} algorithm into an anytime algorithm.

\paragraph{Elements of regret analysis.}
For a sequence $(T_i)_{i\in\N}$, with $T_i\in\N$ for all $i$, we denote $T_{-1} = 0$,
and $T_0$ is always taken non-zero, $T_0 > 0$ (\ie, $T_0\in\N^*$).
We only consider \emph{non-decreasing} and \emph{diverging} sequences
(that is, $\forall i,\; T_{i+1} \geq T_i$, and $T_i \to \infty$ for $i \to \infty$).


\begin{definition}[Last Term $L_T$]\label{def:lastTerm}

    For a non-decreasing diverging sequence $(T_i)_{i\in\N}$
    and $T\in\N$,
    we can define $L_T\bigl((T_i)_{i\in\N}\bigr)$ by
    \begin{equation}
        \forall T\geq1,\;\;\;\;
        L_T\bigl((T_i)_{i\in\N}\bigr) := \min\left\{ i \in \N : T_i > T \right\}.
    \end{equation}

    It is simply denoted $L_T$ when there is no ambiguity (\eg, if the doubling sequence is chosen).
\end{definition}


$\DTr(\cA)$ reinitializes its underlying algorithm $\cA$ at each time $T_i$,
and in generality the total regret is upper bounded by the regret on
each sequence $\{T_i,\dots, T_{i+1} - 1\}$.
By considering the last partial sequence $\{T_{L_T-1}, \dots, T - 1\}$, this splitting can be used to get
a generic upper bound~\eqref{eq:decompositionIneq_UB} by taking into account a larger last sequence (up to $T_{L_T} - 1$).
And for stochastic bandit models, the \iid{} hypothesis on the rewards streams makes the splitting on each sequence an equality, so we can also get the lower bound \eqref{eq:decompositionIneq_LB} by excluding the last partial sequence.
Lemma~\ref{lem:decomposition} is
proved in Appendix~\ref{sub:proof_decompositionDTr}.



\begin{lemma}[Regret Lower and Upper Bounds for $\DTr$]\label{lem:decomposition}

    For \emph{any bandit model} and algorithm $\cA$ and horizon $T$, one has the generic upper bound
    \vspace*{-5pt}  
    \begin{align}
        R_T(\DTr(\cA, (T_i)_{i\in\N}))
        \leq \sum_{i=0}^{L_T} R_{T_i - T_{i-1}}(\cA_{T_i - T_{i-1}}). \tag{LB}\label{eq:decompositionIneq_LB}
    \end{align}

    Under a \emph{stochastic bandit model}, one has furthermore the lower bound
    \begin{align}
        R_T(\DTr(\cA, (T_i)_{i\in\N}))
        \geq \sum_{i=0}^{L_{T-1}} R_{T_i - T_{i-1}}(\cA_{T_i - T_{i-1}}). \tag{UB}\label{eq:decompositionIneq_UB}
    \end{align}
\end{lemma}

As expected, the key to obtain regret guarantees for a Doubling Trick algorithm is to carefully choose the doubling sequence $(T_i)_{i\in\N}$.
Empirically, one can verify that sequences with slow growth gives terrible results, and for example using an arithmetic progression typically gives a linear regret.
%
Building on this result, we will prove that if $\cA$ satisfies a certain regret bound ($R_T = \cO(T^{\gamma})$, $\cO((\log T)^{\delta})$, or $\cO(T^{\gamma} (\log T)^{\delta})$) then an appropriate anytime version of $\cA$ with a certain doubling trick can conserve the regret bound, with an explicit constant multiplicative loss $\ell > 1$.
In this paper, we study in depth two families of sequences: first geometric and then exponential growths.

\section{What the Geometric Doubling Trick Can and Can't Do}\label{sec:geometricDoublingTrick}

We define geometric doubling sequences, and prove that they can be used
to conserve bounds in $\bigO{T^{\gamma}(\log T)^{\delta}}$ with $\gamma>0$ but cannot be used to conserve bounds in $\bigO{(\log T)^{\delta}}$.


\begin{definition}[Geometric Growth]\label{def:geomSequence}
    For $b\in\R$, $b>1$ and $T_0\in\N^*$,
    the sequence defined by $T_i = \lfloor T_0 b^i \rfloor$
    is non-decreasing and diverging, and satisfies
    \vspace*{-10pt}  
    \begin{align}
        \forall T < T_0,\;
        L_{T} = 0,
        \;\;\text{and}\;\;
        \forall T \geq T_0,\;
        L_T = \left\lceil \log_b\left( \frac{T}{T_0} \right) \right\rceil, \label{eq:geomSequence2} \\
        \forall i > 0,\;\;
        T_0 (b - 1) b^{i-1} - 1
        \leq T_i - T_{i-1}
        \leq 1 + T_0 (b - 1) b^{i-1}. \label{eq:geomSequence1}
    \end{align}
    Asymptotically for $i$ and $T\to\infty$,
    $T_i = \bigO{b^i}$ and
    $L_T \sim \log_b(T) = \bigO{\log T}$.
\end{definition}



\subsection{Conserving a Regret Upper Bound with Geometric Horizons}

A geometric doubling sequence allows to conserve a minimax bound (\ie, $R_T = \mathcal{O}(\sqrt{T})$).
It was suggested, for instance, in~\cite[Ex.2.9]{CesaLugosi06}.
We generalize this result in the following theorem, proved in Appendix~\ref{sub:proof_keep45}, by extending it from $\sqrt{T}$ bounds to bounds of the form $T^{\gamma} (\log T)^{\delta}$ for \emph{any} $0<\gamma<1$ and $\delta \geq 0$.
Note that no distinction is done on the case $\delta=0$ neither in the expression of the constant loss, nor in the proof.

\begin{theorem}\label{thm:keep45}
    If an algorithm $\cA$ satisfies
    $ R_T(\cA_T) \leq c \; T^{\gamma} (\log T)^{\delta} + f(T)$,
    for $0<\gamma<1$, $\delta \geq 0$
    and for $c > 0$, and an increasing function $f(t) = \smallO{t^{\gamma} (\log t)^{\delta}}$ (at $t\to\infty$),
    then the anytime version $\cA':=\DTr(\cA, (T_i)_{i\in\N})$ with the geometric sequence $(T_i)_{i\in\N}$ of parameters $T_0\in\N^*,b>1$ (\ie, $T_i = \lfloor T_0 b^i\rfloor$)
    \emph{with the condition} $T_0 (b-1) > 1$ if $\delta>0$,
    satisfies,
    \begin{equation}\label{eq:keep45_1}
        R_T(\cA') \leq \ell(\gamma, \delta, T_0, b) \; c \; T^{\gamma} \left(\log T\right)^{\delta} + g(T),
    \end{equation}
    with a increasing function $g(t) = \smallO{t^{\gamma} (\log t)^{\delta}}$,
    and a \emph{constant loss} $\ell(\gamma, \delta, T_0, b) > 1$,
    \begin{equation}\label{eq:keep45_2}
        \ell(\gamma, \delta, T_0, b) := \left(\frac{\log(T_0(b-1) + 1)}{\log(T_0(b-1))}\right)^{\delta} \; \times \; \frac{b^{\gamma} (b-1)^{\gamma}}{b^{\gamma} - 1}.
    \end{equation}
\end{theorem}

For a fixed $\gamma$ and $\delta$, minimizing $\ell(\gamma, \delta, T_0, b)$ does not always give a unique solution.
On the one hand, if $\gamma \gtrsim 0.384$, there is a unique solution $b^*(\gamma) > 1$ minimizing the $\frac{b^{\gamma} (b-1)^{\gamma}}{b^{\gamma} - 1}$ term, solution of $b^{\gamma+1} - 2b + 1 = 0$, but without a closed form if $\gamma$ is unknown.
On the other hand, for any $\gamma$, the term depending on $\delta$ tends quickly to $1$ when $T_0$ increases.

\paragraph{Practical considerations.}
Empirically, when $\gamma$ and $\delta$ are fixed and known, there is no need to minimize $\ell$ jointly.
It can be minimized separately by first minimizing $\frac{b^{\gamma} (b-1)^{\gamma}}{b^{\gamma} - 1}$,
that is by solving $b^{\gamma+1} - 2b + 1 = 0$ numerically (\eg, with Newton's method), and then by taking $T_0$ large enough so that the other term is close enough to $1$.

For the usual case of $\gamma=1/2$ and $\delta=0$ (\ie, bounds in $\sqrt{T}$), the optimal choice of $b$ is $\frac{3+\sqrt{5}}{2}$ leading to $\ell \simeq 3.33$,
and the usual choice of $b=2$ gives $\ell \simeq 3.41$ (see Corollary~\ref{cor:keep4} in appendix).
Any large enough $T_0$ gives similar performance, and empirically $T_0 \gg K$ is preferred, as most algorithms explore each arm once in their first steps
(\eg, $T_0 = 200$ for $K=9$ for our experiments).

\subsection{A Regret Lower Bound with Geometric Horizons}\label{sec:lowerboundsGeom}

We observe that the constant loss in Eq.~\eqref{eq:keep45_2} from the previous Theorem~\ref{thm:keep45} blows up when $\gamma$ goes to zero,
giving the intuition that no geometric doubling trick could be used to preserve a logarithmic bound (\ie, with $\gamma=0$, $\delta>0$).
This is confirmed by the lower bound given below.

\begin{theorem}\label{thm:LowerBoundGeom_log}
    For \emph{stochastic models},
    if $\cA$ satisfies
    $ R_T(\cA_T) \geq c \; (\log T)^{\delta}$,
    for $c > 0$ and $\delta>0$,
    then the anytime version $\cA':=\DTr(\cA, (T_i)_{i\in\N})$ with the geometric sequence $(T_i)_{i\in\N}$ of parameters $T_0\in\N^*,b>1$ (\ie, $T_i = \lfloor T_0 b^i \rfloor$) satisfies this lower bound for a certain constant $c' > 0$,
    \begin{equation}\label{eq:LowerBoundGeom_log_1}
        \forall T \geq 1,
        L_T \geq 2 \implies
        R_T(\cA') \geq
        c' \; \left(\log T\right)^{\delta+1}.
    \end{equation}
    %
    This implies that $R_T(\cA') = \Omega((\log T)^{\delta+1})$, which proves that a geometric sequence \emph{cannot} be used to conserve a logarithmic regret bound.
\end{theorem}

Theorem~\ref{thm:LowerBoundGeom_log} implies that a geometric sequence \emph{cannot} be used to conserve a finite-horizon lower bound like $R_T(\cA_T) \geq c \log(T)$. A complementary lower bound, stated as Theorem~\ref{thm:LowerBoundGeom_sqrt} in Appendix~\ref{sub:LowerBoundGeom_sqrt}, shows that if the regret is lower bounded at finite horizon by $R_T(\cA_T) \geq c \sqrt{T}$, then a comparable lower bound for the Doubling Trick algorithm $\DTr(\cA)$, possibly with a larger constant.

This special case ($\delta=1$) is indeed the most interesting, as in the stochastic case the regret of any uniformly efficient algorithm is at least logarithmic (\cite{LaiRobbins85}), and efficient algorithm with logarithmic regret have been exhibited.
%
%
%
If $R_T(\cA_T)/{\log T}$ is bounded, then using a geometric sequence in the doubling trick algorithm is a bad idea as it guarantees a blow up in the regret, that is $R_T(\DTr(\cA, (T_i)_{i\in\N})) = \Omega((\log T)^2)$.
This result is the reason we need to consider successive horizons growing faster than a geometric sequence (\ie, such that $\log(T_i) \gg i$),
like the exponential sequence, which is studied in Section~\ref{sec:exponentialDoublingTrick}.

\subsection{Proof of Theorem~\ref{thm:LowerBoundGeom_log}}\label{proof:LowerBoundGeom_log}

    Let $\cA' := \DTr(\cA, (T_i)_{i\in\N})$ and consider a fixed \emph{stochastic} bandit problem.
    The lower bound \eqref{eq:decompositionIneq_LB} from Lemma~\ref{lem:decomposition} gives
    \begin{align*}
        R_T(\cA')
        &\geq \sum_{i=0}^{L_T-1} R_{T_i - T_{i-1}}(\cA_{T = T_i - T_{i-1}}) \\
        \intertext{
            We bound $T_i - T_{i-1} \geq T_0 (b - 1) b^{i-1} - 1$ for any $i>0$, thanks to Definition~\ref{def:geomSequence},
            and we can use the hypothesis on $\cA$ for each regret term.
        }
        &\geq \sum_{i=0}^{L_T-1} c (\log(T_i - T_{i-1}))^{\delta}
        \geq c \sum_{i=1}^{L_T-1} \bigl(\log\left(T_0 (b-1) b^{i-1} - 1\right)\bigr)^{\delta} \\
        &= c \sum_{i=0}^{L_T-2} \bigl(\log\left(T_0 (b-1) b^i - 1\right)\bigr)^{\delta} \;\;\;\;\;\;(\text{with } i := i - 1)\\
        \intertext{
            Let $x_i := T_0 (b-1) b^i > 0$.
            If we have $T_0 (b-1) > 1$ (see below ($\clubsuit$) in Page~\pageref{eq:clubsuit} for a discussion on the other case),
            then Lemma~\ref{lem:logxm1_logx} (Eq.~\eqref{eq:logxm1_logx_2}) gives
            $\log(x_i - 1) \geq \frac{\log(T_0 (b-1) - 1)}{\log(T_0 (b-1))} \log(x_i)$ as $x_i > 1$.
            For lower bounds, there is no need to handle the constants tightly, and we have $x_i \geq b^i$ by hypothesis, so let call this constant $c' = c \left(\frac{\log(T_0 (b-1) - 1)}{\log(T_0 (b-1))}\right)^{\delta} > 0$, and thus it simplifies to
        }
        &\geq c' \sum_{i=0}^{L_T-2} \left(\log(b^i)\right)^{\delta} \\
        \intertext{
            A sum-integral minoration for the increasing function $t \mapsto t^{\delta}$ (as $\delta>0$)
            gives $\sum_{i=0}^{L_T-2} \left(\log(b^i)\right)^{\delta} = (\log b)^{\delta} \sum_{i=1}^{L_T-2} i^{\delta} \geq (\log b)^{\delta} \int_{0}^{L_T-2} t^{\delta} \dt = \frac{(\log b)^{\delta}}{\delta+1} \left(L_T-2\right)^{\delta+1}$ (if $L_T \geq 2$), and so
        }
        R_T(\cA') &\geq c' \frac{(\log b)^{\delta}}{\delta+1} (L_T-2)^{\delta+1} \\
        \intertext{
            For the geometric sequence, we know that $L_T \geq \log_b\left(\frac{T}{T_0}\right) \geq \log_b(T)$,
            and $\log_b(T) - 2 \sim \log_b(T)$ at $T\to\infty$ so there exists a constant $0 < c'' < 1$ such that $L_T - 2 \geq c'' \log_b(T)$ for $T$ large enough ($\geq \frac{2}{b-1}$), and such that $L_T \geq 2$.
            And thus we just proved that there is a constant $c''' > 0$ such that
        }
        R_T(\cA') &\geq c''' (\log T)^{\delta+1} =: g(T).
    \end{align*}
    So this proves that for $T$ large enough, $R_T(\cA') \geq g(T)$ with $g(T) = \bigO{(\log T)^{\delta+1}}$,
    and so $R_T(\cA') = \Omega((\log T)^{\delta+1})$,
    which also implies that $R_T(\cA')$ \emph{cannot} be a $\bigO{(\log T)^{\delta}}$.

    \label{eq:clubsuit}
    ($\clubsuit$) If we do not have the hypothesis $T_0(b-1) > 1$, the same proof could be done,
    by observing that from $i \geq i_0$ large enough, we have $x_i \geq b^{i-i_0}$ (as soon as $b^{i_0} \geq \frac{1}{T_0 (b-1)} > 0$, \ie, $i_0 \geq \lceil -\log_b(T_0 (b-1)) \rceil \geq 1$),
    and so the same arguments can be used, to obtain a sum from $i = i_0 + 1$ instead of from $i = 1$.
    For a fixed $i_0$, we also have $L_T - 2 - i_0 \geq c'' \log(T)$ for a (small enough) constant $c''$, and thus we obtain the same result.
    \hfill{}
    $\blacksquare$

\section{What Can the Exponential Doubling Trick Do?}\label{sec:exponentialDoublingTrick}

We define exponential doubling sequences, and prove that they can be used to conserve bounds in $\bigO{(\log T)^{\delta}}$, unlike the previously studied geometric sequences.
Furthermore, we provide elements showing that they may also conserve bounds in $\bigO{T^{\gamma}}$ or $\bigO{T^{\gamma}(\log T)^{\delta}}$.


\begin{definition}[Exponential Growth]\label{def:expSequence}
    For $a,b\in\R$, $a,b>1$ and $T_0\in\N^*$,
    if $\tau := \frac{T_0}{a} \in \R, \geq 0$,
    then the sequence defined by
    $T_i := \lfloor \tau a^{b^{i}} \rfloor$
    is non-decreasing and diverging, and satisfies
    \begin{equation}\label{eq:expSequence}
        \forall T < T_0,\;
        L_{T} = 0,
        \;\;\text{and}\;\;
        \forall T \geq T_0,\;
        L_T = \left\lceil \log_b\left( \log_a\left( \frac{T}{\tau}\right)\right) \right\rceil.
    \end{equation}
    Asymptotically for $i$ and $T\to\infty$,
    $T_i = \bigO{a^{b^i}}$ and
    $L_T \sim \log_b(\log_a(\frac{T}{\tau})) = \bigO{\log \log T}$.
\end{definition}



\subsection{Conserving a Regret Upper Bound with Exponential Horizons}

An exponential doubling sequence allows to conserve a problem-dependent bound on regret (\ie, $R_T = \bigO{\log T}$). This was already used in particular cases by ~\cite{Auer10} and more recently by \cite{Liauetal2017}.
We generalize this result in the following theorem.



\begin{theorem}\label{thm:keep52}
    If an algorithm $\cA$ satisfies
    $ R_T(\cA_T) \leq c \; T^{\gamma} (\log T)^{\delta}+ f(T)$,
    for $0\leq\gamma<1$, $\delta \geq 0$,
    and for $c > 0$, and an increasing function $f(t) = \smallO{t^{\gamma} (\log t)^{\delta}}$ (at $t\to\infty$),
    then the anytime version $\cA':=\DTr(\cA, (T_i)_{i\in\N})$ with the exponential sequence $(T_i)_{i\in\N}$ of parameters $T_0\in\N^*,a,b>1$ (\ie, $T_i = \lfloor \frac{T_0}{a} a^{b^i}\rfloor$),
    satisfies the following inequality,
    \begin{equation}\label{eq:keep52_1}
        R_T(\cA') \leq \ell(\gamma, \delta, T_0, a, b) \; c \left(T^b\right)^{\gamma} \left(\log T\right)^{\delta} + g(T).
    \end{equation}
    with an increasing function $g(t) = \smallO{(t^b)^{\gamma} (\log t)^{\delta}}$,
    and a \emph{constant loss} $\ell(\gamma, \delta, T_0, a, b) > 0$,
    \begin{equation}\label{eq:keep52_2}
        \ell(\gamma, \delta, T_0, a, b) :=
        \begin{cases}
            \left(\frac{a}{T_0}\right)^{(b-1)\gamma} \; \frac{b^{2\delta}}{b^{\delta} - 1} > 0
            & \text{ if } \delta > 0 \\
            1 + \frac{1}{(\log(a)) (\log(b^{\gamma}))} > 1
            & \text{ if } \delta = 0
        \end{cases}
    \end{equation}
\end{theorem}

This result first shows that an exponential doubling trick can preserve a logarithmic regret bound ($O(\log(T))$, which corresponds to $\gamma=0$ and $\delta=1$),
with a multiplicative constant loss $\ell \geq 4$.
It can further be applied to bounds of the generic form $R_T = \bigO{T^{\gamma} (\log T)^{\delta}}$, but with a \emph{significant loss} as $T^{\gamma}$ becomes $T^{b\gamma}$, additionally to the constant multiplicative loss $\ell > 0$.
However, it is important to notice that for $\gamma>0$, the constant $\ell$ can be made arbitrarily small
%
(with a large enough first step $T_0$).
This observation is encouraging,
and let the authors think that a tighter upper bound could be proved.
%

\begin{remark}\label{cor:Keep5}
%
An interesting particular case of Theorem~\ref{thm:keep52} is the following ($\gamma =0, \delta=1$ and $f(t) = 0$).
        \begin{equation}
        R_T(\cA_T) \leq c \; \log(T)
        \implies
        R_T(\DTr(\cA, (\lfloor T_0^{b^i} \rfloor)_{i\in\N})) \leq \frac{b^2}{b-1} \; c \; \log(T).
    \end{equation}
    In this upper bound, the optimal choice of $b$ is $b=2$, which yields a constant multiplicative loss of $\ell(\gamma=0, b) = 4$. It can be observed that this loss is twice smaller as the loss of $8.0625$ obtained by~\cite[Sec.4]{Auer10}\footnote{In \cite{Auer10}, the authors obtained a loss of $258/32 = 8.0625 \geq 8$, as the ratio between the constants for the $\log(T)$ terms, respectively $258$ in Th.4.1 and $32$ in Th.3.1.}.
\end{remark}


\subsection{A Regret Lower Bound with Exponential Horizons}\label{sec:lowerboundsExpo}


Assuming the upper bound of Theorem~\ref{thm:keep52} obtained for $\gamma > 0$ are tight would lead to think that exponential doubling tricks cannot preserve minimax regret bounds of the form $O(\sqrt{T}))$. If true, such a conjecture would need to be supported by a lower bound (a counterpart of Theorem~\ref{thm:LowerBoundGeom_log}). Theorem~\ref{thm:LowerBoundExpo_sqrt} provides such a lower bound, but as discussed below, combined with Theorem~\ref{thm:keep52}, this result rather advocates the use of exponential doubling tricks.  Theorem~\ref{thm:LowerBoundExpo_sqrt} is proved in Appendix~\ref{sub:proof_LowerBoundExpo_sqrt}.

%

\begin{theorem}\label{thm:LowerBoundExpo_sqrt}
    For \emph{stochastic models},
    if $\cA$ satisfies
    $ R_T(\cA_T) \geq c \; T^{\gamma}$,
    for $c > 0$ and $0 < \gamma \leq 1$,
    then the anytime version $\cA':=\DTr(\cA, (T_i)_{i\in\N})$ with the exponential sequence $(T_i)_{i\in\N}$ of parameters $T_0\in\N^*$, $a>1$, $b>1$ (\ie, $T_i = \lfloor \frac{T_0}{a} a^{b^i} \rfloor$), satisfies this lower bound for a certain constant $c' > 0$,
    \begin{equation}
        \forall T \geq 1,\;\;\;\;
        R_T(\cA') \geq c' \; \left(T^{\frac{1}{b}}\right)^{\gamma}.
    \end{equation}
\end{theorem}


We already saw that any exponential doubling trick can conserve logarithmic problem-dependent regret bounds.
If we could take $b\to1$ in the two previous Theorems~\ref{thm:keep52} and \ref{thm:LowerBoundExpo_sqrt},
both results would match and prove that there exists an exponential doubling trick that can also be used to conserve minimax regret bounds.
This argument is not so easy to formulate, as $b$ cannot depend on $T$, but it supports our belief that exponential doubling tricks are good candidates for (asymptotically) preserving both problem-dependent and minimax regret bounds.



\subsection{Proof of Theorem~\ref{thm:keep52}} \label{proof:Keep52}

    Let $\cA' := \DTr(\cA, (T_i)_{i\in\N})$, and consider a fixed bandit problem.
    We first consider the harder case of $\delta>0$, see below in Page~\pageref{eq:spadesuit} in $(\spadesuit)$ for the other case.
    The lower bound \eqref{eq:decompositionIneq_LB} from Lemma~\ref{lem:decomposition} gives
    \begin{align*}
        R_T(\cA')
        &\leq \sum_{i=0}^{L_T} R_{T_i - T_{i-1}}(\cA_{T = T_i - T_{i-1}}) \\
        \intertext{
            We bound naively\footnotemark{} $T_i - T_{i-1} \leq T_i \leq \frac{T_0}{a} a^{b^i}$,
            and we can use the hypothesis on $\cA$ for each regret term,
            as $f$ and $t \mapsto c t^{\gamma} (\log t)^{\delta}$ are non-decreasing for $t\geq 1$ (by hypothesis for $f$ and by Lemma~\ref{lem:tgammalogtdeltaIncreasing}).
        }
        &\leq \sum_{i=0}^{L_T} f(T_i) + c \sum_{i=0}^{L_T} \left(T_i\right)^{\gamma} \left(\log\left(T_i\right)\right)^{\delta}
        \leq g_1(T) + c \sum_{i=0}^{L_T} \left(\frac{T_0}{a} a^{b^i}\right)^{\gamma} \left(\log\left(\frac{T_0}{a} a^{b^i}\right)\right)^{\delta} \\
        \intertext{
            The first part is denoted $g_1(T):= \sum_{i=0}^{L_T} f(T_i)$ and is dealt with Lemma~\ref{lem:ControlLemmaSumOfSmallO}:
            the sum of $f(T_i)$ is a $\smallO{\sum_{i=0}^{L_T} T_i^{\gamma} (\log(T_i))^{\delta}}$, as $f(t) = \smallO{t^{\gamma} (\log t)^{\delta}}$ by hypothesis, and this sum of $T_i^{\gamma} (\log(T_i))^{\delta}$ is proved below to be bounded by $T^{b\gamma} (\log(T))^{\delta}$.
            So $g_1(T) = \smallO{T^{b\gamma} (\log T)^{\delta}}$.
            %
            The second part is $c \left(\frac{T_0}{a}\right)^{\gamma} \sum_{i=0}^{L_T} \left(a^{b^i}\right)^{\gamma} \left(\log\left(\frac{T_0}{a} a^{b^i}\right)\right)^{\delta}$.
            Define $\log^+(x) := \max(\log(x), 0) \geq 0$, so whether $\frac{T_0}{a} \leq 1$ or $ >1$, we always have $\log\left(\frac{T_0}{a} a^{b^i}\right) \leq \log^+\left(\frac{T_0}{a}\right) + \log\left(a^{b^i}\right)$.
            Then we can use Lemma~\ref{lem:expoInequality} (Eq.~\eqref{eq:expoInequality1}) to distribute the power on $\delta$ (as it is $<1$). So $\left(\log\left(\frac{T_0}{a} a^{b^i}\right)\right)^{\delta} \leq \left(\log^+\left(\frac{T_0}{a}\right)\right)^{\delta} + \left(\log(a)\right)^{\delta} \left(b^i\right)^{\delta}$ with the convention that $0^{\delta} = 0$ (even if $\delta=0$),
            and so this gives
        }
        &\leq g_1(T) + c \left(\frac{T_0}{a}\right)^{\gamma} \left[
            \left(\log^+\left(\frac{T_0}{a}\right)\right)^{\delta} \sum_{i=0}^{L_T} (a^{b^i})^{\gamma}
            + \left(\log(a)\right)^{\delta} \sum_{i=0}^{L_T} (a^{b^i})^{\gamma} \left(b^i\right)^{\delta}
        \right] \\
        \intertext{
            %
            If $\gamma=0$ then the first sum is just $L_T+1 = \bigO{\log(\log(T))}$ which can be included in $g_1(T) = \smallO{(\log T)^{\delta}}$ (still increasing),
            and so only the second sum has to be bounded, and a geometric sum gives $\sum_{i=0}^{L_T} \left(b^i\right)^{\delta} \leq \frac{b^{\delta}}{b^{\delta} - 1} (b^{L_T})^{\delta}$.
            But if $\gamma > 0$,
            we can naively
            bound the first sum by $\sum_{i=0}^{L_T} (a^{b^i})^{\gamma} \leq (L_T+1) (a^{b^{L_T}})^{\gamma}$
            Observe that $a^{b^{L_T}} = (a^{b^{L_T}-1})^b \leq (a\frac{T}{T_0})^b$.
            So $a^{b^{L_T}} = \bigO{T^b}$ and $L_T + 1 = \bigO{\log(\log(T))}$, thus the first sum is a $\bigO{T^{b\gamma} \log(\log(T))} = \smallO{T^{b\gamma} (\log T)^{\delta}}$ (as $\delta>0$).
            In both cases, the first sum can be included in $g_2(T)$ which is still a $\smallO{T^{b\gamma} (\log T)^{\delta}}$
            Another geometric sum bounds the second sum by $\sum_{i=0}^{L_T} (a^{b^i})^{\gamma} \left(b^i\right)^{\delta} \leq (a^{b^{L_T}})^{\gamma} \sum_{i=0}^{L_T} \left(b^i\right)^{\delta} \leq \frac{b^{\delta}}{b^{\delta} - 1}  (b^{L_T})^{\delta}$.
        }
        &\leq g_1(T)
        + c_1 (a^{b^{L_T}})^{\gamma} (b^{L_T})^{\delta}
        \intertext{
            We identify a constant multiplicative loss
            $c_1 := c \left(\frac{T_0}{a}\right)^{\gamma} \frac{b^{\delta}}{b^{\delta} - 1} \left(\log a\right)^{\delta} > 0$.
            %
            %
            The only term left which depends on $L_T$ is $(a^{b^{L_T}})^{\gamma} (b^{L_T})^{\delta}$,
            and it can be bounded by using
            $b^{L_T} = b b^{L_T-1} \leq b \log_a(a \frac{T}{T_0}) = b + b \log^+_a(\frac{T}{T_0}) \leq b + b \log_a(T)$ (as $T\geq1$),
            and
            again with $a^{b^{L_T}} \leq (a\frac{T}{T_0})^b$.
            The constant part of $b^{L_T}$ also gives a $\bigO{T^{b\gamma}}$ term, that can be included in $g(T) := g_2(T) + (a \frac{T}{T_0})^{b\gamma}$ which is still a $\smallO{T^{b\gamma} (\log T)^{\delta}}$, and is still increasing as sum of increasing functions.
            So we can focus on the last term, and we obtain
        }
        &\leq g(T) + c_1 \left(\frac{b}{\log(a)}\right)^{\delta} \left[ \left(\frac{a}{T_0}\right)^{b} \; T^{b} \right]^{\gamma} \left(\log T\right)^{\delta} \\
        \implies R_T(\cA') &\leq g(T) + \ell(\gamma, \delta, T_0, a, b) \; c T^{b\gamma} \left(\log T\right)^{\delta} \text{ with an increasing } g(t) = \smallO{t^{b\gamma} \left(\log t\right)^{\delta}}.
    \end{align*}
    \footnotetext{Here, using the more subtle bound $T_i - T_i \leq \frac{T_0}{a} a^{b^{i-1}} (\alpha^{b^{i-1}}) + 1$, with $\alpha=a^{b-1}$, from Definition~\ref{def:expSequence}, does not seem to help as it becomes too complicated to handle clearly in the $\log$ terms.}

    So the constant multiplicative loss $\ell$ depends on $\gamma$ and $\delta$ as well as on $T_0$, $a$ and $b$ and is
    \begin{align}
        \ell(\gamma, \delta, T_0, a, b) := \left(\frac{a}{T_0}\right)^{(b-1)\gamma} \;\times\; \frac{b^{2\delta}}{b^{\delta} - 1} > 0,
        \;\;\;\;\text{if}\; \delta > 0.
    \end{align}

    If $T_0 = a$, the loss
    $\ell(\gamma, \delta, T_0, a, b)$ is minimal at $b^*(\delta) = 2^{1/\delta} > 1$
    and for a minimal value of $\min_{b > 1} \ell(\gamma, \delta, T_0, a, b) = 4$ (\textbf{for any} $\delta$ and $\gamma$).
    Finally, the $a/T_0$ part tends to $0$ if $T_0\to\infty$ so the loss can be made as small as we want, simply by choosing a $T_0$ large enough (but constant \emph{w.r.t.} $T$).

    $(\spadesuit)$ Now for the other case of $\delta=0$, we can start similarly,
    but instead of bounding naively
    $\sum_{i=0}^{L_T} (a^{b^i})^{\gamma}$
    by $(L_T+1) (a^{b^{L_T}})^{\gamma}$
    we use Lemma~\ref{lem:DoubleExpSumIneq}
    to get a more subtle bound:
    $\sum_{i=0}^{L_T} (a^{b^i})^{\gamma} \leq a^{\gamma} + (1 + \frac{1}{(\log(a)) (\log(b^{\gamma}))}) (a^{b^{L_T}})^{\gamma}$.
    The constant term gets included in $g(T)$, and for the non-constant part,
    $(a^{b^{L_T}})^{\gamma}$ is handled similarly.
    Finally we obtain the loss
    \begin{align}\label{eq:spadesuit}
        \ell(\gamma, 0, T_0, a, b) := 1 + \frac{1}{(\log(a)) (\log(b^{\gamma}))} > 1.
    \end{align}

\vspace*{-10pt}  
\hfill{}
$\blacksquare$

\section{Numerical Experiments}\label{sec:experiments}

We illustrate here the practical cost of using Doubling Trick, for two interesting non-anytime algorithms that have recently been proposed in the literature:
\emph{Approximated Finite-Horizon Gittins indexes}, that we refer to as \AFHG, by \cite{Lattimore16a} (for Gaussian bandits with known variance)
and \KLUCBpp{} by \cite{Menard17} (for Bernoulli bandits).

We first provide some details on these two algorithms, and then illustrate the behavior of Doubling Tricks applied to these algorithms with different doubling sequences.

\subsection{Two Index-Based Algorithms}\label{sub:twoIndexBasedAlgorithms}

%
We denote by
$X_k(t) := \sum_{s < t} Y_{A(s), s} \mathbbm{1}(A(s) = k)$ the accumulated rewards from arm $k$,
and $N_k(t) := \sum_{s < t} \mathbbm{1}(A(s) = k)$ the number of times arm $k$ was sampled.
Both algorithms $\cA$ assume to know the horizon $T$.
They compute an index $I_k^{\cA}(t)\in\R$ for each arm $k\in\{1,\dots,K\}$ at each time step $t\in\{1,\dots,T\}$,
and use the indexes to choose the arm with highest index, \ie, $A(t) := \arg\max_{k\in\{1,\dots,K\}} I_k(t)$ (ties are broken uniformly at random).

\begin{itemize}
    \item The algorithm \AFHG{} can be applied for Gaussian bandits with variance $V$ ($=1$ for our experiments).
    Let $m(T, t) = T - t + 1 \geq 1$, and let
    \begin{equation}\label{eq:index_AFHG}
        I_k^{\AFHG}(t) := \frac{X_k(t)}{N_k(t)} + \sqrt{\frac{V}{N_k(t)} \log\left( \frac{m(T, t)}{N_k(t) \log^{1/2}\left( \frac{m(T, t)}{N_k(t)} \right)} \right)}.
    \end{equation}

    \item The algorithm \KLUCBpp{} can be applied for bounded rewards in $[0,1]$, and in particular for Bernoulli bandits.
    The binary Kullback-Leibler divergence is $\kl(x,y) := x \log(x/y) + (1-x) \log((1-x)/(1-y))$ (for $0<x,y<1$),
    and let $\log^+(x) := \max(0, \log(x)) \geq 0$.
    Let the function $g(n, T) := \log^+\left(\frac{T}{K n} (1 + \left(\log^+(\frac{T}{K n})\right)^2)\right)$ for $n \leq T$,
    and finally let
    \begin{equation}\label{eq:index_KLUCBpp}
        I_k^{\KLUCBpp}(t) := \sup\limits_{q \in [0, 1]} \left\{ q : \kl\left(\frac{X_k(t)}{N_k(t)}, q\right) \leq \frac{g(N_k(t), T)}{N_k(t)} \right\}.
    \end{equation}
\end{itemize}


\subsection{Experiments}\label{sub:experiments}



\begin{figure}[H]
    \centering
    \includegraphics[width=0.95\textwidth]{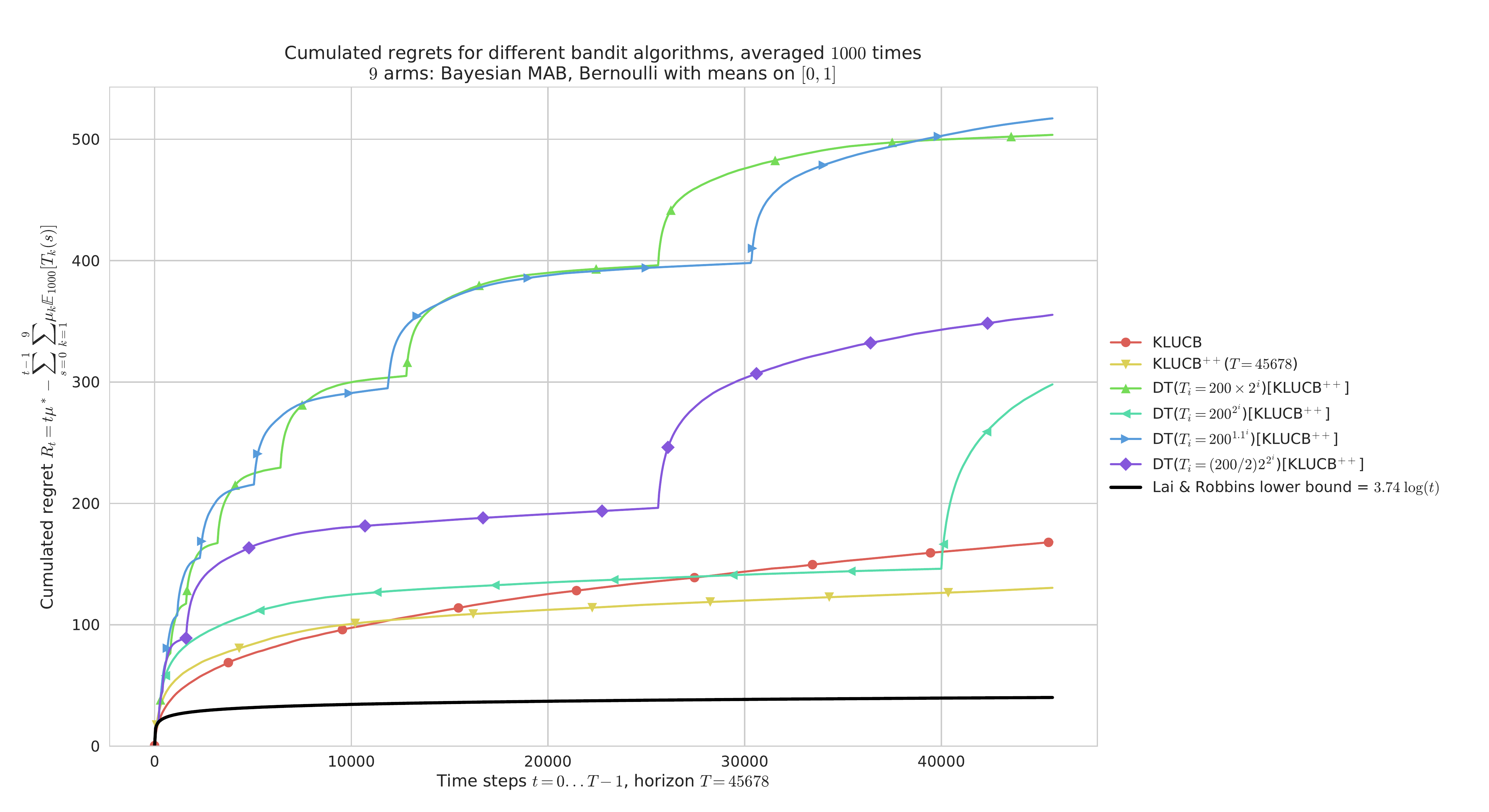}
    \caption{Regret for \DT, for $K=9$ Bernoulli arms, horizon $T=45678$, $n=1000$ repetitions and $\boldsymbol{\mu}$ taken uniformly in $[0,1]^K$. Geometric doubling ($b=2$) and slow exponential doubling ($b=1.1$) are too slow, and short first sequences make the regret blow up in the beginning of the experiment. At $t=40000$ we see clearly the effect of a new sequence for the best doubling trick ($T_i = 200 \times 2^i$). As expected, \KLUCBpp{} outperforms \KLUCB, and if the doubling sequence is growing fast enough then $\DT(\KLUCBpp)$ can perform as well as \KLUCBpp{} (see for $t < 40000$).}
    \label{fig:bernoulliBandits_DoublingTrick_Restart_BayesianProblem}
\end{figure}

\begin{figure}[H]
    \centering
    \includegraphics[width=0.93\textwidth]{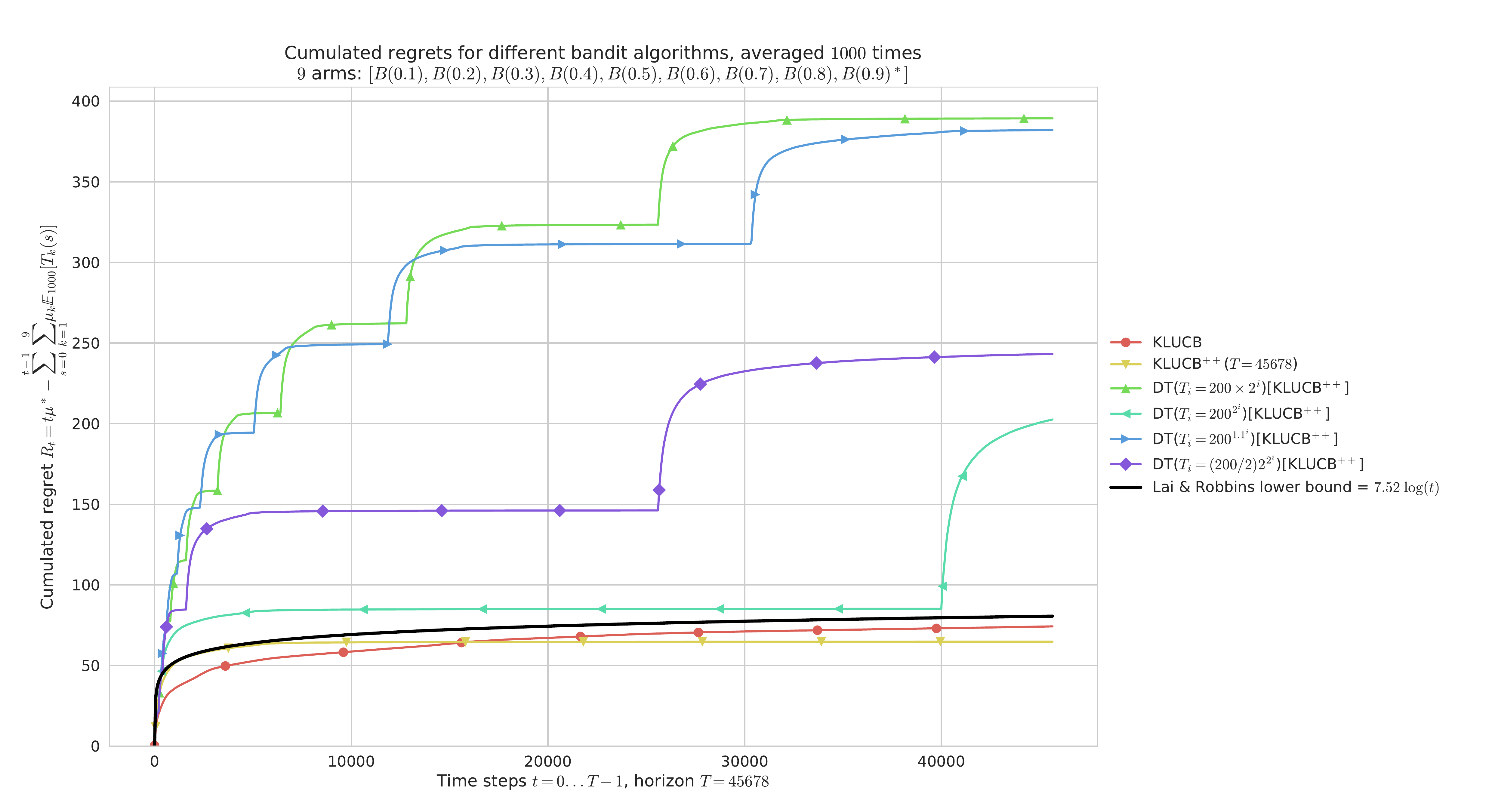}
    \caption{Similarly but for $\boldsymbol{\mu}$ evenly spaced in $[0,1]^K$ ($\{0.1,\dots,0.9\}$). Both \KLUCB{} and \KLUCBpp{} are very efficient on ``easy'' problems like this one, and we can check visually that they match the lower bound from \cite{LaiRobbins85}. As before we check that slow doubling are too slow to give reasonable performance.}
    \label{fig:bernoulliBandits_DoublingTrick_Restart_fixedProblem}
    \vspace*{-15pt}  
\end{figure}

\begin{figure}[!h]
    \centering
    \includegraphics[width=0.93\textwidth]{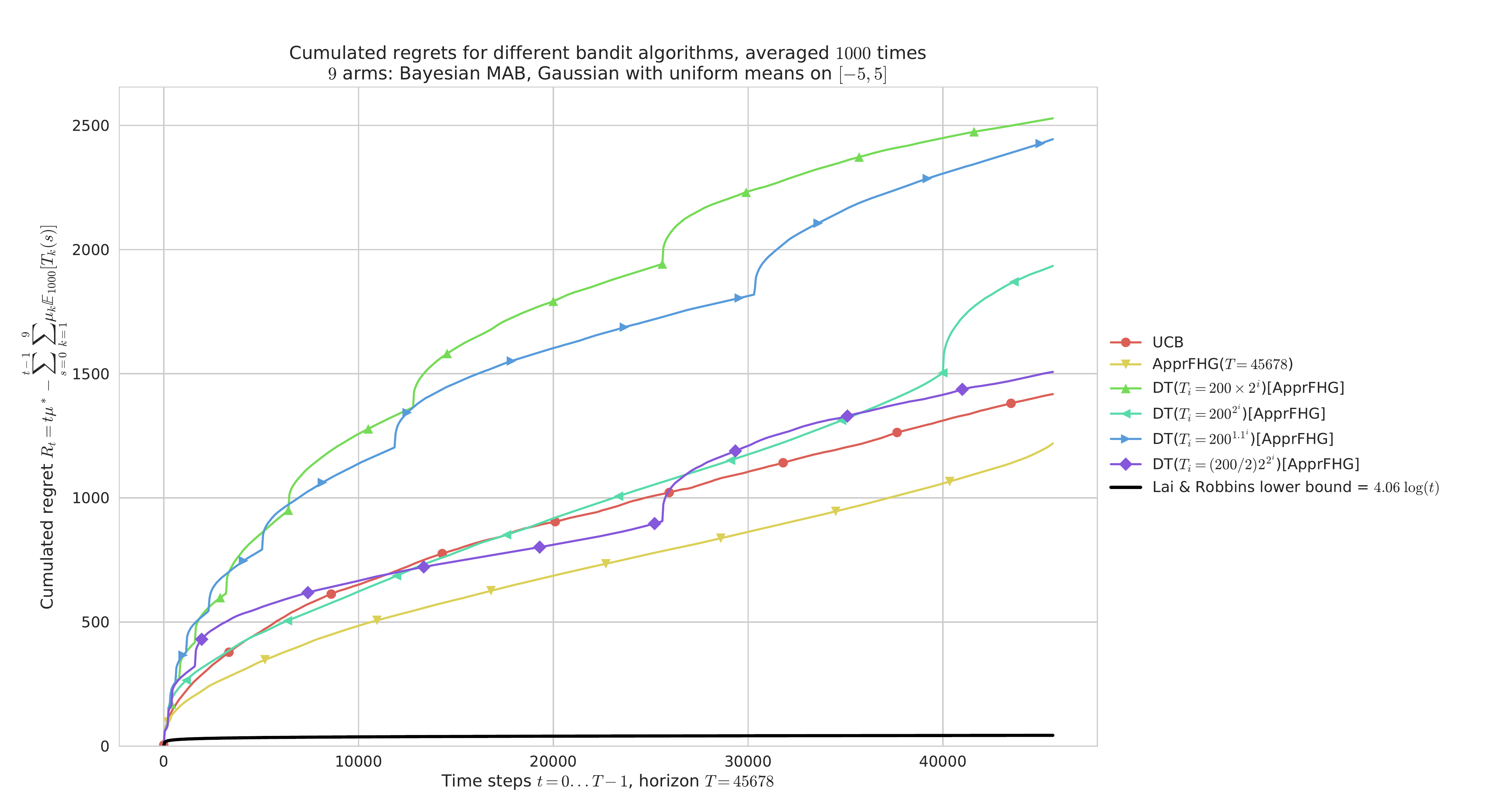}
    \caption{Regret for $K=9$ Gaussian arms $\cN(\mu, 1)$, horizon $T=45678$, $n=1000$ repetitions and $\boldsymbol{\mu}$ taken uniformly in $[-5,5]^K$ and variance $V=1$. On ``hard'' problems like this one, both \UCB{} and \AFHG{} perform similarly and poorly \emph{w.r.t.} to the lower bound from \cite{LaiRobbins85}. As before we check that geometric doubling ($b=2$) and slow exponential doubling ($b=1.1$) are too slow, but a fast enough doubling sequence does give reasonable performance for the anytime \AFHG{} obtained by Doubling Trick.}
    \label{fig:gaussianBandits_DoublingTrick_Restart_fixedProblem}
\end{figure}

We present some results from numerical experiments on Bernoulli and Gaussian bandits. More results are presented in Appendix~\ref{sec:otherExperiments}. 
%
We present in pages~\pageref{fig:bernoulliBandits_DoublingTrick_Restart_BayesianProblem}
and \pageref{fig:bernoulliBandits_DoublingTrick_Restart_fixedProblem}
results for $K=9$ arms
and horizon $T=45678$ (to ensure that no choice of sequence were lucky and had $T_{L_T-1} = T$ or too close to it).
We ran $n=1000$ repetitions of the random experiment, either on the same  ``easy'' bandit problem $\bmu$ (with evenly spaced means),
or on $n$ different random instances $\bmu$ sampled uniformly in $[0,1]^K$,
and we plot the average regret on $n$ simulations.
The black line without markers is the (asymptotic) lower bound in $\sum_{k \neq k^*} (\kl(\mu_k, \mu^*))^{-1} \log T$, from \cite{LaiRobbins85}.
We consider
\KLUCBpp{} for Bernoulli bandits (Figures~\ref{fig:bernoulliBandits_DoublingTrick_Restart_BayesianProblem}, \ref{fig:bernoulliBandits_DoublingTrick_Restart_fixedProblem})
or \AFHG{} for Gaussian bandits (Figures~\ref{fig:gaussianBandits_DoublingTrick_Restart_fixedProblem}, \ref{fig:gaussianBandits_DoublingTrick_Restart_fixedProblem2}),

Each doubling trick algorithm uses the same $T_0=200$ as a first guess for the horizon.
%
We include both the non-anytime version that knows the horizon $T$, and different anytime versions to compare the choice of doubling trick.
To compare against an algorithm that does not need the horizon,
we also include \KLUCB{}~\citep{KLUCBJournal} as a baseline for Bernoulli bandits and for Gaussian bandits (in the Gaussian version, the divergence used is $\mathrm{kl}(x,y)=(x-y)^2/2$, and the algorithm is referred to as \UCB{}).
We consider are a geometric doubling sequence with $b=2$, and two different exponential doubling sequences: the ``classical'' $b=2$ and a ``slower'' one with $b=1.1$. Both use $a=T_0=200$, and the last one is using $a=2$, $b=2$.
Despite what was proved theoretically in Theorem~\ref{thm:keep52}, using $a=T_0$ and a large enough $T_0$ improves regarding to using $a=2$ and a leading $(T_0/a)$ factor.



Another version of the Doubling Trick with ``no restart'', denoted $\DTnr$, is presented in Appendix~\ref{sec:DTnr}, but it is only an heuristic and cannot be applied to any algorithm $\cA$.
Algorithm~\ref{algo:DTnr} can be applied to \KLUCBpp{} or \AFHG{} for instance, as they use $T$ just as a numerical parameter (see Eqs.~\ref{eq:index_AFHG} and \ref{eq:index_KLUCBpp}), but its first limitation is that it cannot be applied to \DMEDp{} \citep{Honda10} or \Exppp{} \citep{Seldin17}, or any algorithms based on arm eliminations, for example.
A second limitation is the difficulty to analyze this ``no restart'' variant,
due to the unpredictable effect on regret of giving non-uniform prior information to the underlying algorithm $\cA$ on each successive sequence.
An interesting future work would be to analyze it, either in general or for a specific algorithm like \KLUCBpp.
Despite its limitations, this heuristic exhibits as expected better empirical performance than \DT, as can be observed in Appendix~\ref{sec:otherExperiments}.

\section{Conclusion}\label{sec:conclusion}

We formalized and studied the well-known ``Doubling Trick'' for generic multi-armed bandit problems, that is used to automatically obtain an anytime algorithm from any non-anytime algorithm. Our results are summarized in Table~\ref{fig:array_conclusion}. We show that a geometric doubling can be used to conserve minimax regret bounds (in $\sqrt{T}$), with a constant loss (typically $\geq 3.33$),
but cannot be used to conserve problem-dependent bounds (in $\log T$),
for which a faster doubling sequence is needed.
An exponential doubling sequence can conserve logarithmic regret bounds also with a constant loss,
but it is still an open question to know if minimax bounds can be obtained for this faster growing sequence.
Partial results of both a lower and an upper bound, for bounds of the generic form $T^{\gamma} (\log T)^{\delta}$, let use believe in a positive answer.

It is still an open problem to know if an anytime algorithm
can be both asymptotically optimal for the problem-dependent regret (\ie, with the exact constant) and optimal in a minimax regret (\ie, have a $\sqrt{KT}$ regret),
but we believe that using a doubling trick on non-anytime algorithms like \KLUCBpp{} cannot be the solution.
We showed that it cannot work with a geometric doubling sequence, and conjecture that exponential doubling trick would never bring the right constant either.



\newcolumntype{M}[1]{>{\centering\arraybackslash}m{#1}}

\begin{figure}[!t]
    \centering
    \begin{tabular}{M{3cm}|M{5cm}|M{6cm}}
        Bound $\backslash$ Doubling
        & Geometric, $T_i = \lfloor T_0 b^i\rfloor$
        & Exponential, $T_i = \lfloor \frac{T_0}{a} a^{b^i}\rfloor$
        \tabularnewline
        \hline
        \hline
        $(\log T)^{\delta}$
        & \textcolor{red}{$\times$ \emph{Known to fail}}
        \newline
        $R_T(\DT) \geq c' (\log T)^{1+\delta}$ if $R_T(\cA_T) \geq c (\log T)^{\delta}$.
        \newline
        (Theorem~\ref{thm:LowerBoundGeom_log})
        & \textcolor{blue}{$\checkmark$ \emph{Known to work}}, with loss $\ell(\delta, b) = \frac{b^{2\delta}}{b^{\delta}-1} > 1$.
        \newline
        (Theorem~\ref{thm:keep52})
        \tabularnewline
        \hline
        $T^{\gamma}$
        & \textcolor{blue}{$\checkmark$ \emph{Known to work}}, with loss $\ell(\gamma, b) = \frac{b^{\gamma}(b-1)^{\gamma}}{b^{\gamma}-1} > 1$.
        \newline
        (Theorem~\ref{thm:keep45})
        & \textcolor{darkgreen}{? \emph{Partial}}, best known bound is $c_0' (T^{\frac{1}{b}})^{\gamma} \leq R_T (\DTr \leq \ell c (T^b)^{\gamma}$
        with a loss $\ell>1$,
        if $c_0 T^{\gamma} \leq R_T(\cA_T) \leq c T^{\gamma}$.
        \newline
        (Theorems~\ref{thm:keep52}, \ref{thm:LowerBoundExpo_sqrt})
        \tabularnewline
        \hline
        $T^{\gamma} \; (\log T)^{\delta}$
        \newline
        for both
        \newline
        $\gamma>0$, $\delta>0$
        & \textcolor{blue}{$\checkmark$ \emph{Known to work}}, with loss $\ell(\gamma, \delta, T_0, b) = \left(\frac{\log(T_0 (b-1) + 1)}{\log(T_0 (b-1))} \right)^{\delta} \frac{b^{\gamma}(b-1)^{\gamma}}{b^{\gamma}-1} > 1$.
        \newline
        (Theorem~\ref{thm:keep45} )
        & \textcolor{darkgreen}{? \emph{Partial}}, best known bound is $R_T (\DTr) \leq \ell c (T^b)^{\gamma}$ if $R_T(\cA_T) \leq c T^{\gamma}$,
        with a loss $\ell\to0$ for $T_0\to\infty$.
        \newline
        (Theorem~\ref{thm:keep52})
    \end{tabular}
    \caption{Summary of \emph{known \textcolor{blue}{positive $\checkmark$} and \textcolor{red}{negative $\times$} and \textcolor{darkgreen}{partial results ?}}.}
    \label{fig:array_conclusion}
\end{figure}


%
%

%
%
%
%
%


\acks{
    This work is supported
    by the French National Research Agency (ANR), under the project BADASS (grant coded: \texttt{N ANR-16-CE40-0002}),
    by the French Ministry of Higher Education and Research (MENESR) and ENS Paris-Saclay.
    %
}


\bibliography{doublingTrick}

\vskip 0.2in
\hr{}
\emph{Note}:
    the simulation code used for the experiments is using Python 3.
    It is open-sourced at \verb|https://GitHub.com/SMPyBandits/SMPyBandits|
    and fully documented at \newline
    \verb|https://SMPyBandits.GitHub.io|.

\clearpage
\newpage

\appendix

\section{Omitted Proofs}\label{sec:missingproofs}

We include here the proofs omitted in the main document.

\subsection{Proof of Lemma~\ref{lem:decomposition}, ``Regret Lower and Upper Bounds for \DTr''}\label{sub:proof_decompositionDTr}

    Let $\cA'$ denote $\DTr(\cA, (T_i)_{i\in\N})$. For every $k\in \{1,\dots, K\}$,
    \begin{align*}
        & \E\left[\sum_{t=1}^T (X_{k,t} - X_{A(t),t})\right] = \sum_{i=0}^{L_T -1} \E\left[\sum_{t=T_{i-1}}^{T_i}(X_{k,t} - X_{A(t),t})\right] +  \E\left[ \sum_{t=T_{L_T-1}}^{T}(X_{k,t} - X_{A(t),t})\right] \\
        & \hspace{0.7cm} \leq
        \sum_{i=0}^{L_T -1} \max_{k \in \{1,\dots,K\}}\E\left[\sum_{t=T_{i-1}}^{T_i} (X_{k,t} - X_{A(t),t})\right] + \max_{k \in \{1,\dots,K\}}\E\left[ \sum_{t=T_{L_T-1}}^{T}(X_{k,t} - X_{A(t),t})\right] \\
        & \hspace{0.7cm} \leq
        \sum_{i=0}^{L_T -1}R_{T_i-T_{i-1}}\left(\cA'\right) + R_{T - T_{L_T-1}}\left(\cA'\right).
    \end{align*}

    Thus, by definition of the regret
    \begin{align*}
        R_T(\cA')
        &\leq
        \sum_{i=0}^{L_T-1} R_{T_i - T_{i-1}}(\cA')+ R_{T - T_{L_T-1}}(\cA') \\
        &=
        \sum_{i=0}^{L_T-1} R_{T_i - T_{i-1}}\left(\cA_{T_{i} - T_{i-1}}\right)
        + R_{ \underbrace{T - T_{L_T-1}}_{\leq T_{L_T} - T_{L_T-1}} }\left(\cA_{T_{L_T} - T_{L_{T-1}}}\right) \\
        &\leq
        \sum_{i=0}^{L_T} R_{T_i - T_{i-1}}\left(\cA_{T_{i} - T_{i-1}}\right).
    \end{align*}

    In the stochastic case, it is well known that the regret can be rewritten in the following way, introducing $\mu_k$ the mean of arm $k$ and $\mu^*$ the mean of the best arm:
    \begin{align*}
        R_T(\cA') &=
        \E\left[\sum_{t=1}^{T}(\mu^* - \mu_{A(t)})\right] \\
        &=
        \sum_{i=0}^{L_{T-1}} \E\left[\sum_{t=T_{i-1}}^{T_i}(\mu^* - \mu_{A(t)})\right] + \E\left[\sum_{t=T_{L_T-1}}^{T}(\mu^* - \mu_{A(t)})\right] \\
        &=
        \sum_{i=0}^{L_T-1} R_{T_i - T_{i-1}}\left(\cA_{T_i - T_{i-1}}\right)) + \underbrace{R_{T - T_{L_T-1}}(\cA')}_{\geq 0}.
    \end{align*}
    and the lower bound follows.
\hfill{}
$\blacksquare$

\newpage

\subsection{Proof of Theorem~\ref{thm:keep45}, ``Conserving a Regret Upper Bound with Geometric Horizons''}\label{sub:proof_keep45}

It is interesting to note that
the proof is valid for both the easiest case when $\delta=0$ (as it was known in \cite{CesaLugosi06} for $\gamma=1/2$)
and the generic case when $\delta\geq0$,
with no distinction.

As far as the authors know, this result in its generality with $\delta\geq0$ is new.

\vspace*{5pt}  
\begin{proof}\label{proof:Keep45}
    Let $\cA' := \DTr(\cA, (T_i)_{i\in\N})$, and consider a fixed bandit problem.
    The upper bound \eqref{eq:decompositionIneq_UB} from Lemma~\ref{lem:decomposition} gives
    \begin{align*}
        R_T(\cA')
        &\leq \sum_{i=0}^{L_T} R_{T_i - T_{i-1}}(A_{T = T_i - T_{i-1}}) \\
        \intertext{
            We can use the hypothesis on $\cA$ for each regret term,
            as $f$ and $t \mapsto c t^{\gamma} (\log t)^{\delta}$ are non-decreasing for $t\geq 1$ (by hypothesis for $f$ and by Lemma~\ref{lem:tgammalogtdeltaIncreasing}).
        }
        &\leq \sum_{i=0}^{L_T} f(T_i - T_{i-1}) + c T_0^{\gamma} (\log T_0)^{\delta}
        + c \sum_{i=1}^{L_T} \left(T_i - T_{i-1}\right)^{\gamma} \left(\log\left(T_i - T_{i-1}\right)\right)^{\delta}
        \intertext{
            The first part is denoted $g_1(T):= \sum_{i=0}^{L_T} f(T_i - T_{i-1}) + c T_0^{\gamma} (\log T_0)^{\delta}$, it is an increasing function as a sum of increasing functions, and it is dealt with by using Lemma~\ref{lem:ControlLemmaSumOfSmallO}:
            the sum of $f(T_i - T_{i-1})$ is a $\smallO{\sum_{i=0}^{L_T} (T_i - T_{i-1})^{\gamma} (\log(T_i - T_{i-1}))^{\delta}}$, as $f(t) = \smallO{t^{\gamma} (\log t)^{\delta}}$ by hypothesis, and this sum of $(T_i - T_{i-1})^{\gamma} (\log(T_i - T_{i-1}))^{\delta}$ is proved below to be bounded by $c' T^{\gamma} (\log(T))^{\delta}$ for a certain constant $c'>0$,
            which gives $g_1(T) = \smallO{T^{\gamma} (\log T)^{\delta}}$.
            For the second part,
            we bound $T_i - T_{i-1} \leq T_0 (b-1) b^{i-1} + 1$ thanks to Definition~\ref{def:geomSequence}.
            Moreover, as $\gamma<1$
            we can use Lemma~\ref{lem:expoInequality} (Eq.~\eqref{eq:expoInequality1}) to distribute the power on $\gamma$,
            so $(T_0 (b-1) b^{i-1} + 1)^{\gamma} \leq (T_0 (b-1) b^{i-1})^{\gamma} + \mathbbm{1}(\gamma\neq0)$ (indeed if $\gamma=0$ both sides are equal to $1$).
            This gives
        }
        &\leq g_1(T) + c
            (T_0 (b-1))^{\gamma} \sum_{i=1}^{L_T} (b^{i-1})^{\gamma} \left(\log(T_i - T_{i-1})\right)^{\delta}
            + c \mathbbm{1}(\gamma\neq0) \sum_{i=1}^{L_T} \left(\log(T_i - T_{i-1})\right)^{\delta} \\
        \intertext{
            If $\gamma\neq 0$,
            the last sum is bounded by $\sum_i^{L_T} (\log T_i)^{\delta} \leq (\log T_0)^{\delta}(L_T + 1) + (\log b)^{\delta} \sum_i^{L_T} i^{\delta}$ which is a $\bigO{L_T^{\delta+1}} = \bigO{(\log T)^{\delta+1}} = \smallO{T^{\gamma} (\log T)^{\delta}}$ (as $\gamma>0$, thanks to a geometric sum), and so it can be included in $g_2(T) = \smallO{T^{\gamma} (\log T)^{\delta}}$.
            If $\gamma=0$, there is only the first sum.
            We bound again $T_i - T_{i-1} \leq T_0 (b-1) b^{i-1} + 1$
            and use Lemma~\ref{lem:logxm1_logx} to bound $\log(T_0 (b-1) b^{i-1} + 1)$
            by $\frac{\log(T_0 (b-1) + 1)}{\log(T_0 (b-1))} \log(T_0 (b-1) b^{i-1}) $ term (as $T_0 (b-1) > 1$ by hypothesis).
        }
        &\leq g_2(T) + c
            (T_0 (b-1))^{\gamma} \sum_{i=1}^{L_T} (b^{i-1})^{\gamma} \left( \frac{\log(T_0 (b-1) + 1)}{\log(T_0 (b-1))} \log(T_0 (b-1) b^{i-1})\right)^{\delta} \\
        \intertext{
            We split the $\log(T_0(b-1)b^{i-1})$ term in two, and
            once again, the term with $\log(T_0(b-1))$ gives a $\bigO{{b^{L_T-1}}}$ (by a geometric sum), which gets included in $g_3(T) = \smallO{T^{\gamma} (\log T)^{\delta}}$.
            We focus on the fastest term, and we can now rewrite the sum from $i=0$ to $L_T-1$,
        }
        &\leq g_2(T) + c
            (T_0 (b-1))^{\gamma} \left(\log(b) \frac{\log(T_0 (b-1) + 1)}{\log(T_0 (b-1))} \right)^{\delta} \sum_{i=0}^{L_T-1} (b^i)^{\gamma} i^{\delta} \\
        \intertext{
            We naively bound $i^{\delta}$ by $(L_T-1)^{\delta}$,
            and use a geometric sum
            to have
        }
        &\leq g_2(T) + c
            (T_0 (b-1))^{\gamma} \left(\log(b) \frac{\log(T_0 (b-1) + 1)}{\log(T_0 (b-1))} \right)^{\delta} (L_T-1)^{\delta} \frac{b^{\gamma}}{b^{\gamma}-1} (b^{L_T-1})^{\gamma} \\
        \intertext{
            Finally, observe that $L_T-1 \leq \log_b(\frac{T}{T_0}) \leq \log_b(T)$, so the $(\log b)^{\delta}$ term simplifies,
            and observe that $b^{L_T-1} \leq \frac{T}{T_0}$ so the $T_0^{\gamma}$ term also simplifies.
            Thus we get
        }
        &\leq g_2(T) + c
            \left(\frac{\log(T_0 (b-1) + 1)}{\log(T_0 (b-1))} \right)^{\delta} \frac{b^{\gamma} (b-1)^{\gamma}}{b^{\gamma}-1} T^{\gamma} (\log T)^{\delta}.
    \end{align*}

    The constant multiplicative loss $\ell$ depends on $\gamma$ and $\delta$ as well as on $T_0$ and $b$,
    and is
    $\ell(\gamma, \delta, T_0, b) := \left(\frac{\log(T_0 (b-1) + 1)}{\log(T_0 (b-1))} \right)^{\delta} \frac{b^{\gamma} (b-1)^{\gamma}}{b^{\gamma}-1} > 1$.
\end{proof}

\paragraph{Minimizing the constant loss?}
This constant loss has two distinct part,
$\ell(\gamma, \delta, T_0, b) = \ell_1(\delta, T_0, b)$ $\ell_2(\gamma, b)$,
with $\ell_1$ depending on $\delta$, $T_0$ and $b$ (equal to $1$ if $\delta=0$),
and $\ell_2$ depending on $\gamma$ and $b$.
\begin{itemize}
    \item
Minimizing this constant loss $\ell_1(\delta, T_0, b) := \left(\frac{\log(T_0 (b-1) + 1)}{\log(T_0 (b-1))} \right)^{\delta}\geq 1$ is independent of $\delta$ (even if it is $0$).
If we assume $b$ to be fixed,
$\ell_1(\delta, T_0, b) \to 1$ when $T_0\to\infty$.
Moreover, for any $\delta$ and $b>1$,
$\ell_1(\delta, T_0, b)$ goes to $1$ very quickly when $T_0$ is large enough.
For instance, for $\gamma=\delta=\frac{1}{2}$ and $b=\frac{3+\sqrt{5}}{2}$ (see Corollary~\ref{cor:keep4}),
then $\ell_1(\delta, T_0, b) \simeq 1.109$ for $T_0=2$, $\simeq 1.01$ for $T_0=10$ and $\simeq 1.0004$ for $T_0=100$.
    \item
To minimize this constant loss $\ell_2(\gamma, b) := \frac{b^{\gamma} (b - 1)^{\gamma}}{b^{\gamma} - 1} > 1$, we fix $\gamma$ and study $h: b \mapsto \ell_2(\gamma, b)$.
The function $h$ is of class $\cC^1$ on $(1, \infty)$ and $h(b) \to +\infty$ for $b\to 1^+$ and $b\to\infty$,
so $h$ has a (possibly non-unique) global minimum and attains it.
Moreover $h'(b) = \frac{\gamma b^{\gamma-1} (b-1)^{\gamma-1}}{(b^{\gamma}-1)^2} \left( b^{\gamma+1} - 2 b + 1 \right)$ has the sign of $b^{\gamma+1} - 2 b + 1$,
which does not have a constant sign and does not have explicit root(s) for a generic $\gamma$.
However, it is easy to minimizing $\ell_2(\gamma, b)$ for $b$ numerically when $\gamma$ is known and fixed (with, \eg, Newton's method).
\end{itemize}


The result from Theorem~\ref{thm:keep45} of course implies the result from \cite[Ex.2.9]{CesaLugosi06}, in the special case of $\delta=0$ and $\gamma=\frac{1}{2}$ (for minimax bounds),
as stated numerically in the following Corollary~\ref{cor:keep4}.

\begin{corollary}\label{cor:keep4}
    If $\gamma=\frac{1}{2}$ and $\delta=0$,
    the multiplicative loss $\ell(\frac{1}{2}, 0, T_0, b)$ does not depend on $T_0$.
    It is then minimal for $b^*(\frac{1}{2}) = \frac{3 + \sqrt{5}}{2} \simeq 2.62$ and its minimum is $\sqrt{\frac{11 + 5 \sqrt{5}}{2}} \simeq 3.33$.
    Usually $b = 2$ is used, which gives a loss of $\frac{\sqrt{2}}{\sqrt{2} - 1} \simeq 3.41$, close to the optimal value.


    In particular, order-optimal and optimal algorithms for the minimax bound
    have $\gamma=\frac{1}{2}$ and $f(t)=0$,
    for which Theorem~\ref{thm:keep52} gives a simpler bound
    \begin{equation}\label{eq:corkeep4_1}
        R_T(\cA_T) \leq c \sqrt{T}
        \implies
        R_T(\DTr(\cA, (T_0 2^i)_{i\in\N})) \leq \frac{\sqrt{2}}{\sqrt{2} - 1} c \sqrt{T}.
    \end{equation}
\end{corollary}

\subsection{Proof of Theorem~\ref{thm:LowerBoundExpo_sqrt}, ``Minimax Regret Lower Bound with Exponential Horizons''}\label{sub:proof_LowerBoundExpo_sqrt}

\begin{proof}\label{proof:LowerBoundExpo_sqrt}
    Let $\cA' := \DTr(\cA, (T_i)_{i\in\N})$, and consider a fixed \emph{stochastic} bandit problem.
    The lower bound \eqref{eq:decompositionIneq_LB} from Lemma~\ref{lem:decomposition} gives
    \begin{align*}
        R_T(\cA')
        &\geq \sum_{i=0}^{L_T-1} R_{T_i - T_{i-1}}(A_{T = T_i - T_{i-1}}) \\
        \intertext{
            We can use the hypothesis on $\cA$ for each regret term,
            and as $0 < \gamma \leq 1$, we can use Lemma~\ref{lem:expoInequality} (Eq.~\eqref{eq:expoInequality3}) to distribute the power on $\gamma$ to ease the proof and obtain
        }
        &\geq c T_0^{\gamma} + c \sum_{i=1}^{L_T-1} (T_i - T_{i-1})^{\gamma} \\
        &\geq c T_0^{\gamma} + c \sum_{i=1}^{L_T-1} \left(T_i^{\gamma} - T_{i-1}^{\gamma}\right) \;\;\;\;\text{(it is a telescopic sum and simplifies)} \\
        &\geq c T_{L_T-1}^{\gamma} \\
        \intertext{
            Observe that
            $T_{L_T-1} \geq (T_{L_T})^{\frac{1}{b}}$ by definition of the exponential sequence (Def.~\ref{def:expSequence}), and $T_{L_T} \geq T$ (Def.~\ref{def:lastTerm}).
            For the $\log (T_{L_T-1})$ term, we simply have $\log (T_{L_T-1}) \geq \frac{1}{b} \log(T)$ so if $c' = c / b$,
            then we obtain what we want
        }
        R_T(\cA')
        &\geq c' \; T^{\frac{\gamma}{b}}.
    \end{align*}
    This lower bound goes from $R_T(\cA_T) = \Omega{T^{\gamma}}$ to $R_T(\DT(\cA)) = \Omega{T^{\frac{\gamma}{b}}}$,
    and it looks very similar to the upper bound from Theorem~\ref{thm:keep52} where
    $R_T(\DT(\cA)) = \bigO{T^{b\gamma}}$
    was obtained from
    $R_T(\cA_T) = \bigO{T^{\gamma}}$.
\end{proof}

\paragraph{Remark}
It does seem sub-optimal to lower bound $T_{L_T-1}$ like this ($T_{L_T-1} \geq (T_{L_T})^{\frac{1}{b}}$), but we remind that $T$ can be located anywhere in the discrete interval $\{T_{L_T-1}, \dots, T_{L_T} - 1\}$, so in the worst case when $T$ is very close to $T_{L_T}$ (and for large enough $T$), we indeed have $T_{L_T-1}^b \sim T_{L_T}$ and $T_{L_T} \sim T$, so with this approach, the lower bound $T_{L_T-1} \geq T^{\frac{1}{b}}$ cannot be improved.


\hr{}

\section{Minimax Regret Lower Bound with Geometric Horizons}\label{sub:LowerBoundGeom_sqrt}

We include here a last result that partly replies to Theorem~\ref{thm:keep45}.
It is more subtle that the lower bound in Theorem~\ref{thm:LowerBoundGeom_log} but still provides an interesting insight:
if $b$ is not chosen carefully (\ie, if $\ell_0(\gamma,b) > 1$), then the anytime version of $\cA_T$ using a geometric Doubling Trick suffers a non-improvable constant multiplicative loss compared to $\cA_T$.

\begin{theorem}\label{thm:LowerBoundGeom_sqrt}
    For \emph{stochastic models},
    if $\cA$ satisfies
    $ R_T(\cA_T) \geq c \; T^{\gamma}$,
    for $0<\gamma<1$ and $c > 0$,
    then the anytime version $\cA':=\DTr(\cA, (T_i)_{i\in\N})$ with the geometric sequence $(T_i)_{i\in\N}$ of parameters $T_0\in\N^*,b>1$ (\ie, $T_i = \lfloor T_0 b^i \rfloor$) satisfies
    \begin{equation}
        L_T \geq 2 \implies
        R_T(\cA') \geq \ell_0(\gamma, b) \; c \; T^{\gamma} + g_0(T).
    \end{equation}
    with $g_0(t) = \bigO{\log t} = \smallO{t^{\gamma}}$,
    and a \emph{constant loss} $\ell_0(\gamma, b)$
    depending only on $\gamma$ and $b$,
    \begin{align}
        \ell_0(\gamma, b) &= \frac{(b-1)^{\gamma}}{b^{\gamma} (b^{\gamma} - 1)} > 0.
    \end{align}
    $\ell_0(\gamma, b)$ is always $> 0$ and tends to $0$ for $b\to\infty$,
    and some choice of $b$ gives $\ell_0(\gamma, b) > 1$.
\end{theorem}
\begin{proof}\label{proof:LowerBoundGeom_sqrt}
    Let $\cA' := \DTr(\cA, (T_i)_{i\in\N})$, and consider a fixed \emph{stochastic} bandit problem.
    Assume $L_T \geq 2$.
    The lower bound \eqref{eq:decompositionIneq_LB} from Lemma~\ref{lem:decomposition} gives
    \begin{align*}
        R_T(\cA')
        &\geq \sum_{i=0}^{L_T-1} R_{T_i - T_{i-1}}(A_{T = T_i - T_{i-1}}) \\
        \intertext{
            We bound $T_i - T_{i-1} \geq T_0 (b - 1) b^{i-1} - 1$ for $i>0$, thanks to Definition~\ref{def:geomSequence},
            and we can use the hypothesis on $\cA$ for each regret term.
            Additionally, we have
            $(T_i - T_{i-1})^{\gamma} \geq (T_0 (b - 1) b^{i-1} - 1)^{\gamma} \geq (T_0 (b-1))^{\gamma} (b^{i-1})^{\gamma} - 1$ by Lemma~\ref{lem:expoInequality} (Eq.~\eqref{eq:expoInequality3}, as $b>1$ and $0<\gamma<1$), thus
        }
        &\geq \sum_{i=0}^{L_T-1} c (T_i - T_{i-1})^{\gamma}
        \geq c T_0^{\gamma} + c T_0^{\gamma} (b-1)^{\gamma} \sum_{i=1}^{L_T-1} (b^{i-1})^{\gamma} - c \sum_{i=1}^{L_T-1} 1 \\
        &\geq c T_0^{\gamma} + c T_0^{\gamma} (b-1)^{\gamma} \sum_{i=0}^{L_T-2} (b^i)^{\gamma} - c (L_T - 1) \\
        \intertext{
            We have $\sum_{i=0}^{L_T-1} (b^i)^{\gamma} = \frac{(b^{L_T-1})^{\gamma} - 1}{b^{\gamma} - 1}$ thanks to a geometric sum (with $\gamma>0$) and thus
        }
        &\geq c T_0^{\gamma} + c T_0^{\gamma} (b-1)^{\gamma} \frac{(b^{L_T-1})^{\gamma} - 1}{b^{\gamma} - 1} + c (1 - L_T)
        \intertext{
            Thanks to Definition~\ref{def:geomSequence}, $b^{L_T - 1}$ satisfies $b^{L_T - 1} \geq \frac{1}{b} \frac{T}{T_0}$.
            Let $g_0(T) := c T_0^{\gamma} \left(\frac{(b - 1)^{\gamma}}{b^{\gamma} - 1} - 1\right) + c (L_T - 1) = \bigO{1} + \bigO{\log_b(\frac{T}{T_0})} = \bigO{\log T} = \smallO{T^{\gamma}}$ and $g_0(T) > 0$, then we have
        }
        &\geq c \frac{(b - 1)^{\gamma}}{b^{\gamma}(b^{\gamma} - 1)} T^{\gamma} - \left[ c T_0^{\gamma} \left( \frac{(b - 1)^{\gamma}}{b^{\gamma} - 1} - 1 \right) + c (L_T - 1) \right].
    \end{align*}
    We obtain as announced,
    $R_T(\cA') \geq \ell_0(b) \; c T^{\gamma} + g_0(T)$ with $g_0(T) = \bigO{\log T} = \smallO{T^{\gamma}}$.
\end{proof}

\paragraph{Maximizing the constant loss?}
To maximize\footnote{For the largest possible lower bound, we try to maximize the constant loss in the lower bound.} $\ell_0(\gamma, b) := \frac{(b - 1)^{\gamma}}{b^{\gamma}(b^{\gamma} - 1)} > 0$, we fix $\gamma$ and study the function $h: b \mapsto \ell_0(\gamma, b)$.
The function $h$ is of class $\cC^1$ on $(1, \infty)$ and $h(b) \to +\infty$ for $b\to 1^+$ and $h(b) \to 0$ for $b\to\infty$.
Moreover $h'(b) = - \gamma \frac{(b - 1)^{\gamma - 1}}{b^{\gamma + 1}\left(b^{\gamma} - 1\right)^{2}}  \left(- 2 b^{\gamma} + b^{\gamma + 1} + 1\right)$
has the same sign as $f(b) := - \left(- 2 b^{\gamma} + b^{\gamma + 1} + 1\right)$.
The function $f$ is of class $\cC^1$, with $f(1) = 0$ and  $f'(b) = - (\gamma+1)(b - \frac{2\gamma}{\gamma+1}) b^{\gamma-1}$, and as $0 < \gamma < 1$, $\frac{2\gamma}{\gamma+1} < 1$ so $f'(b) < 0$ for all $b>1$.
Thus $f$ is decreasing, and $\forall b>1, f(b) < f(1) = 0$.
So $h'$ has a negative sign, and this allows to conclude that $h$ is decreasing,
and so $b\mapsto \ell_0(\gamma, b)$ has no global maximum at fixed $\gamma$,
and $\ell_0\to\infty$ if $b \to 1^+$.
%

\paragraph{Relationship with the upper bound.}
For any $b>1$, we compare $\ell_0(\gamma, b)$ with $\ell(\gamma, b)$
and we see that, interestingly, $\ell_0(\gamma, b) = \ell_2(\gamma, b) / b^{2\gamma}$,
with $\ell_2(\gamma, b)$ from Theorem~\ref{thm:keep45}.
%
%
For the particular case of $\gamma=\frac{1}{2}$, this lower bound also leads to another interesting remark:
if $b$ is chosen to minimize the loss in the upper bound (Theorem~\ref{thm:keep45}, $b^*(\frac{1}{2})=\frac{3+\sqrt{5}}{2}$), then this lower bound gives $\ell_0(\frac{1}{2}, b^*(\frac{1}{2})) = 1 + \frac{\sqrt{2}}{2} \simeq 1.71 > 1$, which proves that this choice of geometric doubling trick cannot be used to conserve an optimal algorithm,
\ie, the constant loss cannot be made as close to $1$ as we want.


\section{An Efficient Heuristic, the Doubling Trick with ``No Restart''}\label{sec:DTnr}


Let $\cA' := \DTnr(\cA, (T_i)_{i\in\N})$ denotes the following Algorithm~\ref{algo:DTnr}.
The only difference with $\DTr$ (Algorithm~\ref{algo:DTr}) is that the history from all the steps from $t=1$ to $t=T_i$ is used to reinitialize the new algorithm $\cA^{(i)}$.
To be more precise, this means that a fresh algorithm $\cA^{(i)}$ is created, and then fed with successive observations $(A'(s), Y_{A'(s), s})$ for all $1 \leq s < t$, like if it was playing \emph{from the beginning}.
Note that $A^{(i)}$ could have chosen a different path of actions, but we give it the observations from the previous plays of $\cA'$.

This obviously cannot be applied to any kind of algorithm $\cA$, and for instance any algorithm based on arm elimination (\eg, Explore-Then-Commit approaches like in \cite{GarivierETC2016}) will most surely fail with this approach.
This second doubling trick algorithm $\DTnr$ can be applied in practice if $\cA$ is index-based and uses the horizon $T$ as a simple numerical parameter in its indexes,
like it is the case for instance for
\AFHG{} (cf. Eq.~\eqref{eq:index_AFHG}).
or \KLUCBpp{} (cf. Eq.~\eqref{eq:index_KLUCBpp}).

\vspace*{5pt}  
\begin{algorithm}[H]
        \LinesNumbered  
        \DontPrintSemicolon
        \KwIn{Bandit algorithm $\cA$, and sequence $(T_i)_{i\in\N}$\;}
        \BlankLine
        Let $i = 0$, and initialize algorithm $\cA^{(0)} = \cA_{T_0}$.\\
        \For{$t = 1, \dots, T$}{
            \If(\tcp*[f]{Next horizon $T_{i+1}$ from the sequence}){
                $t > T_i$
            }{
                Let $i = i + 1$.\;
                Initialize algorithm $\cA^{(i)} = A_{T_i - T_{i-1}}$.\;
                Update internal memory of $\cA^{(i)}$ with the history of plays and rewards from $\cA_{i-1}$.
            }
            Play with $\cA^{(i)}$: play arm $A'(t) := A^{(i)}(t - T_i)$, observe reward $r(t) = Y_{A'(t), t}$. \nllabel{line:internalTimer_DTnr}
        }
        \caption{The Non-Restarting Doubling Trick Algorithm, $\cA' = \DTnr(\cA, (T_i)_{i\in\N})$.}
\label{algo:DTnr}
\end{algorithm}
\vspace*{10pt}  


However, it is much harder to state any  theoretical result on this heuristic $\DTnr$.
We conjecture that a regret upper bound similar to \eqref{eq:decompositionIneq_UB} from Lemma~\ref{lem:decomposition} could still be obtained, but it is still an open problem that the authors do not know how to tackle for a generic algorithm.
The intuition is that starting $\cA^{(i)}$ with some previous observations from the (same) \iid{} process $(Y_{k,s})_{k\in\{1,\dots,K\},s\in\N}$ can only improve the performance of $\cA^{(i)}$,
and lead to a smaller regret on the interval $\{T_{i},\dots,T_{i+1}-1\}$.

%

\hr{}

\section{Basic but Useful Results}\label{sec:otherproofs}

All the logarithm $\log$ are taken in base $\e = \exp(1)$ (\ie, natural logarithm $\ln$, but $\log$ is preferred for readability).
Logarithms in a basis $b > 1$ are denoted $\log_b(x) := \frac{\log x}{\log b}$, for any $x\in\R$, $x>0$.

We remind that $\lfloor x \rfloor$ denotes the integer part of $x\in\R$, and for $x>0$, that is
the unique integer $i$ such that $i \leq x < i + 1$.
The only property we use is its definition and the fact that $\lfloor x \rfloor \leq x$.
We also define $\lceil x \rceil := 1 + \lfloor x \rfloor$ for $x \geq 0$, which is the unique integer $j$ such that $j-1 \leq x < j$.

\subsection{Weighted Geometric Inequality}

\begin{lemma}[\emph{Weighted} Geometric Inequality]\label{lem:GeomWeightedSumIneq}
    For any $n \in \N^*$, $b > 1$ and $\delta > 0$,
    and if $f$ is a function of class $\cC^1$, non-decreasing and non-negative on $[0, \infty)$,
    we have
    \begin{equation}
        \sum_{i=0}^{n} f(i) (b^i)^{\delta} \leq
        \frac{b^{\delta}}{b^{\delta} - 1} f(n) (b^n)^{\delta}.
    \end{equation}

\end{lemma}
\begin{proof}\label{proof:ineq2}
    %
    By hypothesis, $f$ is non-decreasing, so $\forall i\in\{0,\dots,n\}, f(i) \leq f(n)$,
    and so by using the sum of a geometric sequence, we have
    \begin{align*}
        \sum_{i=0}^{n} f(i) (b^i)^{\delta}
        \leq f(n) \left( \sum_{i=0}^{n} (b^i)^{\delta} \right)
        \leq f(n) \frac{1}{b^{\delta} - 1} (b^{n+1})^{\delta}
        = \frac{b^{\delta}}{b^{\delta} - 1} \left( f(n) (b^n)^{\delta} \right).
    \end{align*}
    $f(i) = 1$ gives the geometric inequality.
    Note that if we make $\delta\to0$, the left sum converges to $\sum_{i=0}^{n-1} f(i)$ and the right term diverges to $+\infty$, making this inequality completely uninformative.
\end{proof}

\subsection{Another Sum Inequality}

This second result is similar to the previous one but for a ``doubly exponential'' sequence, \ie, $a^{b^i}$,
as it also bounds a sum of increasing terms by a constant times its last term.

\begin{lemma}\label{lem:DoubleExpSumIneq}
    For any $n \in \N^*$, $a > 1$, $b > 1$ and $\gamma > 0$,
    we have
    \begin{equation}
        \sum_{i=0}^{n} (a^{b^i})^{\gamma} \leq
        a^{\gamma} + \left(1 + \frac{1}{(\log(a)) (\log(b^{\gamma}))}\right)
        (a^{b^n})^{\gamma}
        = \bigO{(a^{b^n})^{\gamma}}.
    \end{equation}
\end{lemma}
\begin{proof}\label{proof:DoubleExpSumIneq}
    We first isolate both the first and last term in the sum and focus on the from $i=1$ sum up to $i=n-1$.
    As the function $t \mapsto (a^{b^t})^{\gamma}$ is increasing for $t \geq 1$,
    we use a sum-integral inequality,
    and then the change of variable $u := \gamma b^t$,
    of Jacobian $\dt = \frac{1}{\log b} \frac{\du}{u}$,
    gives
    \begin{align*}
        \sum_{i=1}^{n-1} (a^{b^i})^{\gamma}
        &\leq \int_{1}^{n} a^{\gamma b^t} \dt
        \leq \frac{1}{\log(b^{\gamma})} \int_{\gamma b}^{\gamma b^n} \frac{a^{u}}{u} \du \\
        \intertext{
            Now for $u\geq1$, observe that $\frac{a^{u}}{u} \leq a^{u}$,
            and as $\gamma b > 1$, we have
        }
        &\leq \frac{1}{\log(b^{\gamma})} \int_{\gamma b}^{\gamma b^n} a^{u} \du
        \leq \frac{1}{\log(b^{\gamma})} \frac{1}{\log(a)} a^{\gamma b^n}
        = \frac{1}{(\log(a)) (\log(b^{\gamma}))} {(a^{b^n})}^{\gamma}.
    \end{align*}
    Finally, we obtain as desired,
    $\sum\limits_{i=0}^{n} (a^{b^i})^{\gamma} \leq
    a^{\gamma} + (a^{b^n})^{\gamma} + \frac{1}{(\log(a)) (\log(b^{\gamma}))} {(a^{b^n})}^{\gamma}$.
\end{proof}

\subsection{Basic Functional Inequalities}

These functional inequalities are used in the proof of the main theorems.



\begin{lemma}[Generalized Square Root Inequalities]\label{lem:expoInequality}
    For any $x,y \geq 0$ and $0 < \delta < 1$,
    \begin{equation}\label{eq:expoInequality1}
        (x + y)^{\delta} \leq x^{\delta} + y^{\delta}.
    \end{equation}
    And conversely for any $0 < \delta < 1$, and $x,y \geq 0$, if $x \geq y$ then
    \begin{equation}\label{eq:expoInequality3}
        (x - y)^{\delta} \geq x^{\delta} - y^{\delta}.
    \end{equation}
\end{lemma}
\begin{proof}\label{proof:expoInequality}
    Fix $y\geq0$. Let $f(x):=(x+y)^{\delta} - (x^{\delta} + y^{\delta})$ for $x\geq0$.
    First, $f(0)= y^{\delta} - y^{\delta} = 0$,
    and as $\delta>0$, $f$ is differentiable on $[0, \infty)$, with
    $f'(x) = (\log \delta)(x+y)^{\delta} - (\log \delta)x^{\delta} = (\log \delta)( (x+y)^{\delta} - x^{\delta})$,
    and as $\delta < 1$, $\log\delta < 0$, and $(x+y)^{\delta} \geq x^{\delta}$, so $f'(x) \leq 0$ for any $x\geq0$.
    Therefore, $f$ is non-increasing, and so $\forall x\geq0, f(x) \leq f(0) = 0$, so $f$ is non-positive, giving the desired inequality (for any $y\geq0$ and any $x\geq0$).


    The second inequality is a direct application of the first one.
    Assume $x \geq y$, and let $x' = x - y \geq 0$, then
    $(x' + y)^{\delta} \leq (x')^{\delta} + y^{\delta}$.
    This gives $(x-y)^{\delta} = (x')^{\delta} \leq (x'+y)^{\delta} - y^{\delta} = x^{\delta} - y^{\delta}$.
    %
\end{proof}

\begin{lemma}[Bounding $\log(x-\Delta)$]\label{lem:logxm1_logx}
    Let $x_0 > 1$ and $0 < \Delta < x_0$ (\eg, $\Delta \leq 1$), then
    \begin{equation}\label{eq:logxm1_logx_1}
        \forall x \geq x_0,\;\;
        \frac{\log(x_0-\Delta)}{\log(x_0)} \log(x)
        \leq \log(x-\Delta)
        \leq \log(x).
    \end{equation}
    With $\Delta=1$, it implies that if $T_0>1$, $b>1$ satisfy $T_0(b-1)>1$,
    then for any $i\in\N$, we have
    \begin{equation}\label{eq:logxm1_logx_2}
        \log(T_0(b-1)b^i-1) \geq \frac{\log(T_0(b-1)-1)}{\log(T_0(b-1))} \log(T_0(b-1)b^i).
    \end{equation}
\end{lemma}
\begin{proof}\label{proof:logxm1_logx}
    Let $f(x) := \frac{\log(x-\Delta)}{\log(x)}$, defined for $x\geq x_0$.
    It is of class $\cC^1$, and by differentiating, we have $f'(x) = \frac{\log(x) - \log(x-\Delta)}{x (\log x)^2} > 0$ as $\log$ is increasing.
    So $f$ is increasing, and its minimum is attained at $x = x_0$,
    \ie, $\forall x\geq x_0, f(x) \geq f(x_0) = \frac{\log(x_0 - \Delta)}{\log(x_0)} > 0$, which gives Eq.~\eqref{eq:logxm1_logx_1}.

    The corollary is immediate but stated explicitly for clarity when used in page~\pageref{proof:LowerBoundGeom_log}.
\end{proof}

\begin{lemma}\label{lem:tgammalogtdeltaIncreasing}
    For any $\gamma > 0$ and $\delta>0$,
    the function $f: x \mapsto x^{\gamma}(\log x)^{\delta}$ is increasing on $[1, \infty)$.
\end{lemma}
\begin{proof}\label{proof:tgammalogtdeltaIncreasing}
    $f$ is of class $\cC^1$ on $[1, \infty)$.
    First, if $\gamma > 0$,
    we have $f'(x) = x^{\gamma-1}(\log x)^{\delta-1} \left( \delta + \gamma \log(x)\right)$.
    So $f'(x) > 0$ if and only if $\delta + \gamma \log(x) \geq 0$, that it $x \geq \exp(- \frac{\delta}{\gamma})$.
    But $x \geq 1$ and $ < 0$ so $f'(x)$ is always positive, and thus $f$ is increasing.
    Then, if $\gamma = 0$,
    we have $f'(x) = \delta \frac{1}{x}(\log x)^{\delta-1} > 0$ as $x > 1$ gives $\log x > 0$
    and so $(\log x)^{\delta-1} > 0$.

    It is also true if $\gamma \geq 0$, $\delta \geq 0$ if not both are zero simultaneously.
\end{proof}

\subsection{Controlling an Unbounded Sum of Dominated Terms}

This Lemma is used in the proofs of our upper bounds (Theorems~\ref{thm:keep45} and \ref{thm:keep52}),
to handle the sum of $f(T_i)$ terms.
In particular, it can be applied to $(T_i)$ the geometric sequence and $g(t) = h(t) = t^{\gamma}$ or $g(t) = h(t) = t^{\gamma} (\log t)^{\delta}$ (for Theorem~\ref{thm:keep45})
for $\gamma>0$ and $\delta\geq0$;
or $(T_i)$ the exponential sequence, $g(t) = t^{\gamma} (\log t)^{\delta}$ and $h(t) =  t^{b\gamma} (\log t)^{\delta}$ (for Theorem~\ref{thm:keep52})
for $\gamma\geq0$ and $\delta\geq0$.
Note that it would be obvious if $L_T$ was bounded for $T\to\infty$, but a more careful analysis has to be given as $L_T \to \infty$.

\begin{lemma}\label{lem:ControlLemmaSumOfSmallO}
    Let $f$, $g$ and $h$ be three positive, diverging and non-decreasing functions on $[1, \infty)$, such that
    $f(t) = \smallO{g(t)}$ for $t\to\infty$.
    Let a non-decreasing diverging sequence $(T_i)_{i\in\N}$,
    and a diverging sequence $(L_T)_{T\in\N}$ (\ie, $T_i\to\infty$ for $i\to\infty$ and $L_T\to\infty$ if $T\to\infty)$,
    such that there exists a constant $c\geq0$ satisfying
    $\forall T\geq1, \sum_{i=0}^{L_T} g(T_i) \leq c \times h(T)$.
    Then the (unbounded) sum of dominated terms $f(T_i)$ is still dominated by $h(T)$, \ie,
    \begin{align}\label{eq:ControlLemmaSumOfSmallO}
        f(t) \underset{T\to\infty}{=} \smallO{g(t)}
        \;\text{and}\;
        \exists c\geq0, \sum_{i=0}^{L_T} g(T_i) \underset{\forall T\geq1}{\leq} c \times h(T)
        \implies
        \sum_{i=0}^{L_T} f(T_i) \underset{T\to\infty}{=} \smallO{h(T)}.
    \end{align}
\end{lemma}
\begin{proof}\label{proof:ControlLemmaSumOfSmallO}
    By hypothesis, if $f$ is dominated by $g$, formally $f(t) = \smallO{g(t)}$, then there exists a positive function $\eps(t)$ such that $f(t) = g(t) \eps(t)$ and $\eps(t) \to 0$ for $t\to0$.
    Fix $\eta>0$, as small as we want, then there exists $T_{\eta} \geq 1$ such that
    $\forall t \geq T_{\eta}, \eps(t) \leq \eta$.
    Let $i_{\eta}$ such that $T_{i_{\eta}-1} < T_{\eta} \leq T_{i_{\eta}}$.
    Now for any $T \geq T_{\eta}$ and large enough so that $L_T \geq T_{\eta}$,
    we can start to split the sum
    \begin{align*}
        \sum_{i=0}^{L_T} f(T_i)
        &= \sum_{i=0}^{i_{\eta} - 1} f(T_i) + \sum_{i=i_{\eta}}^{L_T} f(T_i) \\
        \intertext{
            The first sum is naively bounded by $i_{\eta} \times f(T_{i_{\eta} - 1})$ as $f$ is increasing,
            and for the second sum, for any $i\geq i_{\eta}$, $T_i \geq T_{\eta}$ and so $f(T_i) = \eps(T_i) g(T_i) \leq \eta g(T_i)$, thus
        }
        &\leq i_{\eta} f(T_{i_{\eta} - 1}) + \eta \times \left( \sum_{i=i_{\eta}}^{L_T} g(T_i) \right) \\
        \intertext{
            The sum is smaller than a sum on a larger interval, as $g(T_i) \geq 0$ for any $i$,
            and $f$ is increasing so
        }
        &\leq i_{\eta} f(T_{\eta}) + \eta \left( \sum_{i=0}^{L_T} g(T_i) \right) \\
        \intertext{
            But now, $f(T_{\eta}) \leq \eta g(T_{\eta})$ by hypothesis, and this sum is smaller than $c \times h(T)$ also by hypothesis
        }
        &\leq i_{\eta} \eta \times g(T_{\eta}) + \eta c\times  h(T)
        = \eta \left( i_{\eta}  g(T_{\eta}) + c\times  h(T) \right) \\
        \intertext{
            Finally, we use the hypothesis that $h(T)$ is diverging and as $\eta$ and $T_{\eta}$ are both fixed,
            there exists a $\widetilde{T_{\eta}} \geq T_{\eta}$ large enough so that
            $i_{\eta} g(T_{\eta}) \leq h(T)$ for all $T \geq \widetilde{T_{\eta}}$.
            And so we have finally proved that
        }
        &\forall \eta > 0,\;\;
        \exists \widetilde{T_{\eta}} \geq 1,\;\;
        \forall T \geq \widetilde{T_{\eta}},\;\;
        \sum_{i=0}^{L_T} f(T_i) \leq \eta (c+1) \times h(T).
    \end{align*}
    This concludes the proof and shows that $\sum\limits_{i=0}^{L_T} f(T_i) = \smallO{h(T)}$ as wanted.
\end{proof}

\section{Additional Experiments}\label{sec:otherExperiments}

We presents additional experiments, for Gaussian bandits and for the heuristic \DTnr{}.

\subsection{Experiments with Gaussian Bandits (with Known Variance)}\label{sub:experimentsGaussianBandits}

\begin{figure}[!h]
    \centering
    \includegraphics[width=1.00\textwidth]{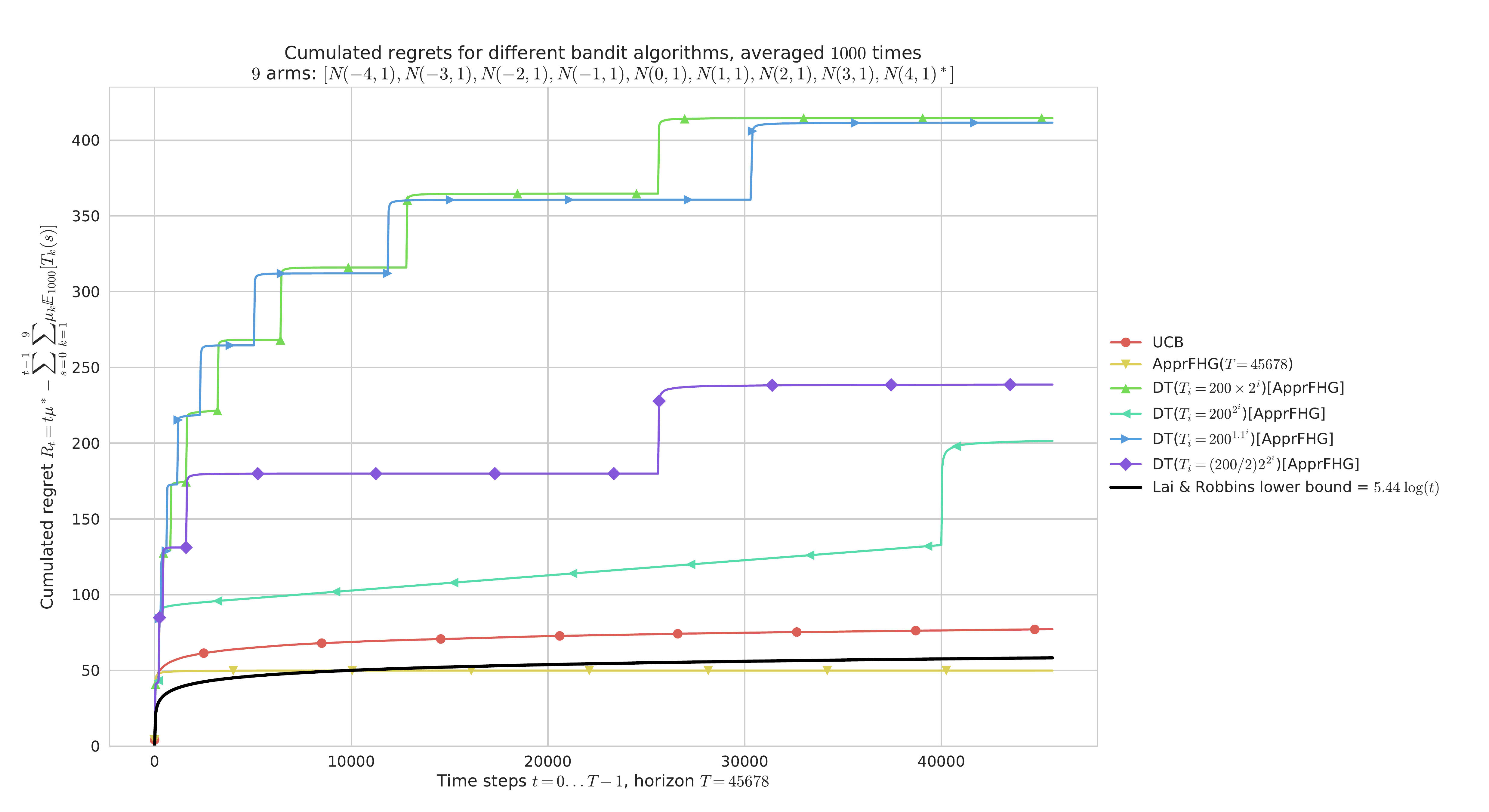}
    \caption{Regret for \DT, for $K=9$ Gaussian arms $\cN(\mu, 1)$, horizon $T=45678$, $n=1000$ repetitions and $\boldsymbol{\mu}$ uniformly spaced in $[-5,5]^K$. On ``easy'' problems like this one, both \UCB{} and \AFHG{} perform similarly and attain near constant regret (identifying the best Gaussian arm is very easy here as they are sufficiently distinct). Each doubling trick also appear to attain near constant regret, but geometric doubling ($b=2$) and slow exponential doubling ($b=1.1$) are slower to converge and thus less efficient.}
    \label{fig:gaussianBandits_DoublingTrick_Restart_fixedProblem2}
\end{figure}

We include here another figure for experiments on Gaussian bandits, see Fig.~\ref{fig:gaussianBandits_DoublingTrick_Restart_fixedProblem2}.

%
%



\subsection{Experiments with \DTnr}\label{sub:experimentsDTnr}

As mentioned previously, the \DTnr{} algorithm (Algorithm~\ref{algo:DTnr})
is only an heuristic so far, as no theoretical guarantee was proved for it.
For the sake of completeness, we also include results from numerical experiments with it, to compare its performance against the ``with restart'' version \DTr.

\begin{figure}[H]
    \centering
    \includegraphics[width=0.92\textwidth]{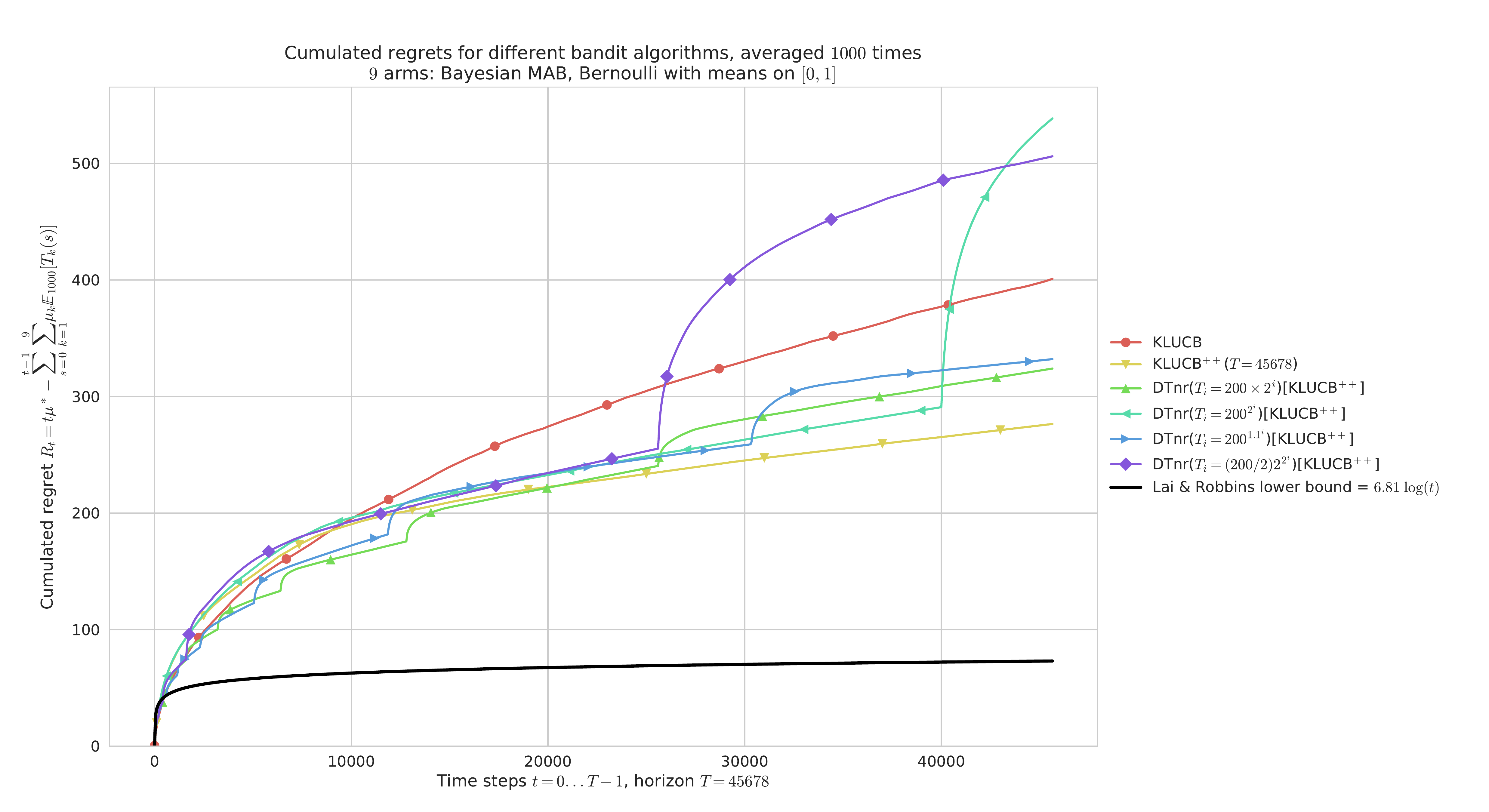}
    \caption{Regret for $K=9$ Bernoulli arms, horizon $T=45678$, $n=1000$ repetitions and $\boldsymbol{\mu}$ taken uniformly in $[0,1]^K$, for \DTnr. Geometric doubling (\eg, $b=2$) and slow exponential doubling (\eg, $b=1.1$) are too slow, and short first sequences make the regret blow up in the beginning of the experiment. At $t=40000$ we see clearly the effect of a new sequence for the best doubling trick ($T_i = 200 \times 2^i$). As expected, \KLUCBpp{} outperforms \KLUCB, and if the doubling sequence is growing fast enough then $\DTnr(\KLUCBpp)$ can perform as well as \KLUCBpp.}
    \label{fig:bernoulliBandits_DoublingTrick_noRestart_BayesianProblem}
    \vspace*{-15pt}  
\end{figure}

\begin{figure}[H]
    \centering
    \includegraphics[width=0.92\textwidth]{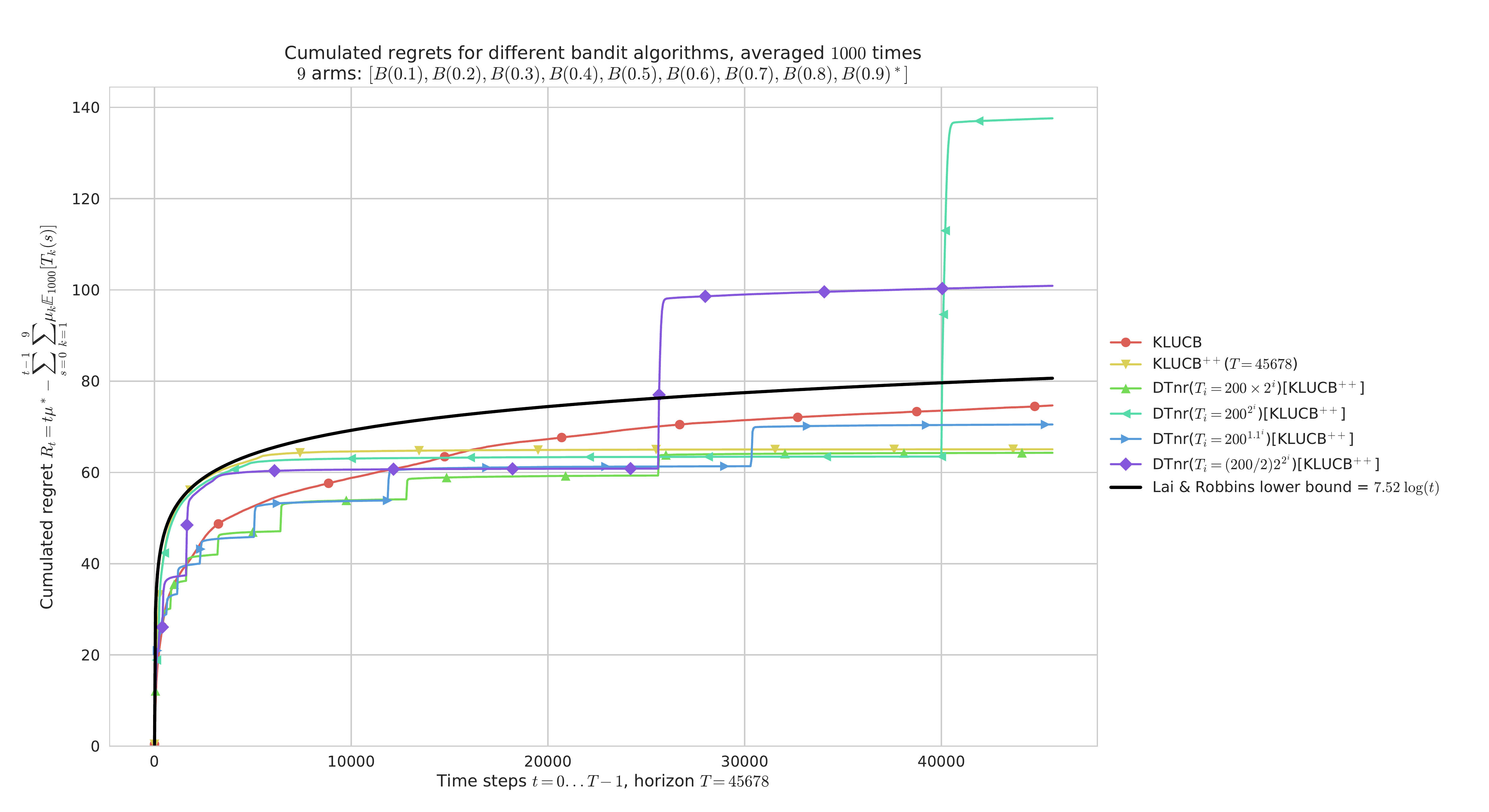}
    \caption{$K=9$ Bernoulli arms with $\boldsymbol{\mu}$ evenly spaced in $[0,1]^K$. On easy problems like this one, both \KLUCB{} and \KLUCBpp{} are very efficient, and here the geometric allows the \DTnr{} anytime version of \KLUCBpp{} to outperform both \KLUCB{} and \KLUCBpp.}
    \label{fig:bernoulliBandits_DoublingTrick_noRestart_fixedProblem}
\end{figure}

As expected, \DTnr{} enjoys much better empirical performance,
and in Figs.~\ref{fig:bernoulliBandits_DoublingTrick_noRestart_BayesianProblem} and \ref{fig:bernoulliBandits_DoublingTrick_noRestart_fixedProblem} we see that a geometric or a slow exponential doubling trick with no restart with \KLUCBpp{} can outperform \KLUCB{} and perform similarly to the non-anytime \KLUCBpp.
But as observed before, the regret blows up after the beginning of each new sequence if the doubling sequence increase too fast (\eg, exponential doubling).
Despite what is proved theoretically in Theorem~\ref{thm:LowerBoundGeom_log}, here we observe that the geometric doubling is the only one to be slow enough to be efficient.

%
%
%
%
%


\end{document}